\documentclass[final]{arxiv}
\usepackage{graphicx}
\usepackage{url}
\usepackage{float}
\restylefloat{table}
\usepackage{lscape}
\usepackage{multirow, pbox}
\usepackage{algorithm}
\usepackage{algorithmic}
\usepackage{amsmath, amsfonts}
\usepackage{graphicx}
\usepackage{caption}
\usepackage{subcaption}
\usepackage{lscape}
\usepackage{longtable}
\usepackage{geometry}
\usepackage{pdflscape}
\usepackage{multirow}

\begin{document}

\title{Twin Neural Networks for the Classification of Large Unbalanced Datasets}
\author{Jayadeva, Himanshu Pant, Sumit Soman and Mayank Sharma\\
Department of Electrical Engineering,\\
Indian Institute of Technology, Delhi, India.\\
\textit{Email: \{jayadeva, eez138524, eez127509, eez142368\}@ee.iitd.ac.in}
\thanks{Prof Jayadeva is with the Department of Electrical Engineering, Indian Institute of Technology, Delhi, India. }%
\thanks{Email: jayadeva@ee.iitd.ac.in}%
}

\begin{abstract}
Twin Support Vector Machines (TWSVMs) have emerged an efficient alternative to Support Vector Machines (SVM) for learning from imbalanced datasets. The TWSVM learns two non-parallel classifying hyperplanes by solving a couple of smaller sized problems. However, it is unsuitable for large datasets, as it involves matrix operations. In this paper, we discuss a Twin Neural Network (Twin NN) architecture for learning from large unbalanced datasets. The Twin NN also learns an optimal feature map, allowing for better discrimination between classes. We also present an extension of this network architecture for multiclass datasets. Results presented in the paper demonstrate that the Twin NN generalizes well and scales well on large unbalanced datasets.
\end{abstract}

\maketitle

\noindent \textbf{Keywords.}
Twin SVM, Neural Network, Unbalanced Datasets, Large Datasets, Large scale learning, Skewed data

\section{Introduction}

Support Vector Machines (SVMs) have been widely used as a machine learning technique for a wide variety of applications. In principle, the SVM aims at finding a maximum-margin hyperplane that best separates samples of two classes. When the two classes are unbalanced or skewed in terms of the number of samples, the SVM is usually biased towards the larger class. The Twin SVM \cite{13} (TWSVM) finds two non-parallel hyperplanes, each of which as as close as possible to points of one class, while being at a distance of at least unity from points of the other class. This involves solving two smaller sized Quadratic Programming Problems (QPPs). Each of the QPPs involves samples of only one of the two classes in the constraints. Consequently, the individual hyperplanes are insensitive to the relative sizes of the two classes. In contrast, a single hyperplane would tend to be biased towards the larger class.

Several extensions of the TWSVM have been proposed in the literature. Reviews of the Twin SVM and its extensions have appeared in AI Review (2014) \cite{twinaireview}, the Annals of Data Science (2014) \cite{twinannalsreview}, and the Egyptian Informatics Journal (2015) \cite{twinegyptianreview}. Some of the extensions include Knowledge based Least Squares TWSVM \cite{33}, Margin Based TWSVM \cite{25}, $\epsilon$-TWSVM \cite{26}, Twin Parametric Margin Classifier \cite{peng2011tpmsvm}, Least-squares Twin Parametric-Margin Classifier \cite{27}, Twin Support Vector Hypersphere Classifier \cite{peng2014twin}, Least-squares Twin Support Vector Hypersphere Classifier \cite{peng2010least}, Structural TWSVM \cite{qi2013structural} among others. The TWSVM has also been used for regression \cite{khemchandani2015twin}, and has proven to be efficient even in the primal formulation \cite{30, 31}. A comprehensive review is also available in \cite{khemchandani2016twin}.
\begin{figure}[hbtp]
\centering
\includegraphics[scale=0.62]{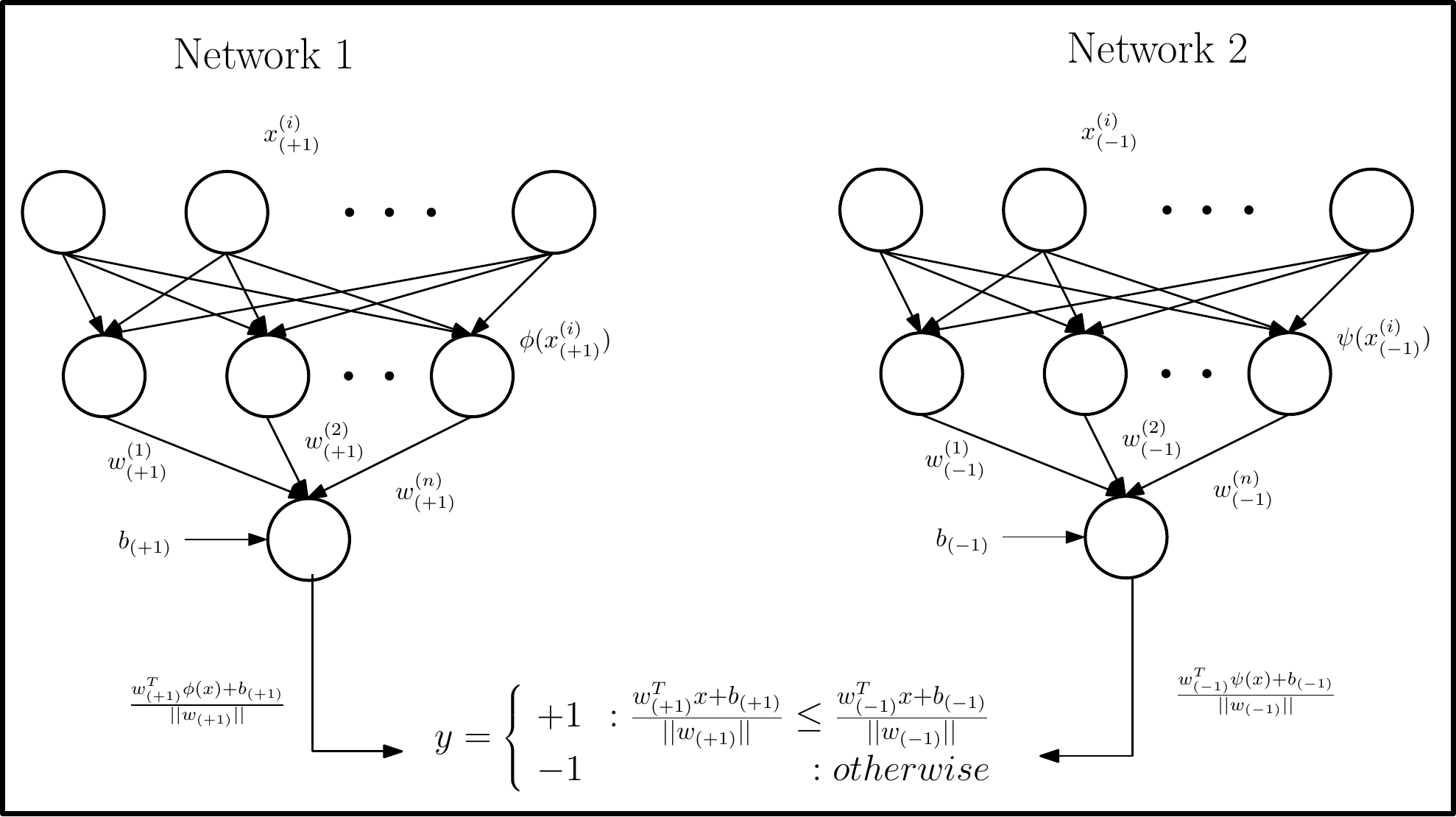}
\caption{Architecture of the Twin Neural Network}
\label{fig:twin_nn}
\end{figure}

However, the TWSVM dual formulation (which is conventionally used) requires the use of several matrix operations including matrix inversion \cite[Eqn (28)-(29)]{13}, which is not practical to implement for large datasets. These operations make it difficult to derive a fast update rule like the SMO or 1SMO \cite{joshi2012using}, and detract from the numerical stability of the algorithm. In this paper, we present a novel neural network architecture called the Twin Neural Network (Twin NN), whose architecture is summarized in Fig. \ref{fig:twin_nn}. The Twin NN is effective in learning from unbalanced datasets, and is significantly faster than the Twin SVM. The neural network also optimizes the feature map, yielding better generalization. It should be emphasized that the Twin NN is not just the extension of the Twin SVM into a neural net framework. Note from Fig. \ref{fig:twin_nn} that the class of a sample is determined by comparing the distances of the sample from the separating hyperplanes defined at the output layer.

\begin{figure}[hbtp]
\centering
\includegraphics[scale=0.8]{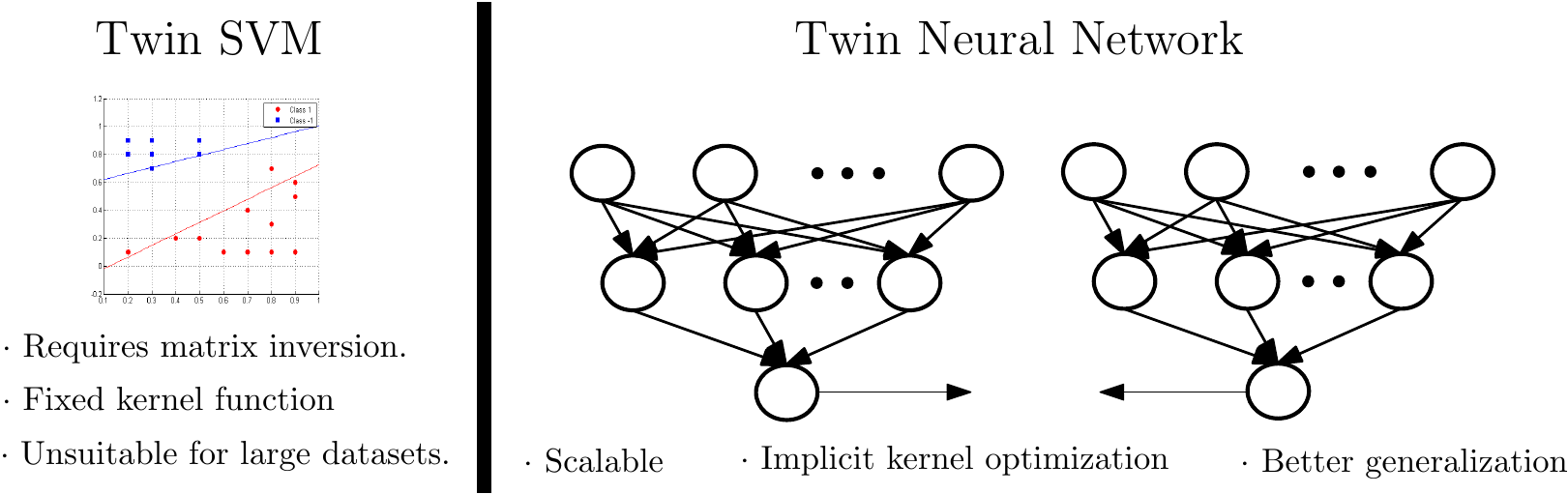}
\caption{Comparing TWSVM and Twin NN}
\label{fig:twin_nn_intro}
\end{figure}

The rest of the paper is organized as follows. Section \ref{sec:twsvm} discusses the formulation for the TWSVM and mentions the dual formulation which is commonly used in practice. Section \ref{sec:twnn} introduces the Twin Neural Network formulation, and its extension for multiclass datasets is discussed in Section \ref{sec:twmc}. This is followed by Section \ref{sec:expts} which shows experimental results to benchmark the performance of the proposed network. Finally, Section \ref{sec:conc} presents the conclusions and future work.

\section{The Twin Support Vector Machine \label{sec:twsvm}}

We begin by motivating the TWSVM from the classical SVM formulation, based on the inability of SVMs to classify unbalanced datasets efficiently. Consider a binary classification dataset of $N$ samples in $M$ dimensions $X=\lbrace x^{(1)}, x^{(2)},..., x^{(N)} | x^{(i)} \in \mathbb{R}^M, \forall i=1,2,..., N\rbrace$, with corresponding labels $Y=\lbrace y^{(1)}, y^{(2)},..., y^{(N)} | y^{(i)} \in \lbrace -1,+1 \rbrace, \forall i=1,2,..., N\rbrace$. We can determine the number of points in each class from this dataset. Let these be denoted as $A=\lbrace  x^{(i)}| y^{(i)}=+1 \rbrace$ for samples of class $(+1)$ and $B=\lbrace  x^{(i)}| y^{(i)}=-1 \rbrace$ for samples of class $(-1)$. The soft-margin Support Vector Machine (SVM) classifier has the following formulation.

\begin{eqnarray}
 \min_{w,b,\xi} && \frac{1}{2} ||w||^2
+ C \sum_{i=1}^l \xi_i \label{primal} \\
\text{subject to,}
&& y^{(i)} (w^T (x^{(i)} + b)) \geq 1 - \xi_i,
\nonumber \\
&& \xi_i \geq 0, i = 1, \ldots, l. \nonumber 
\end{eqnarray}
where the separating hyperplane is given by the vector $[w, b]$, $C$ is the upper bound, $\xi_i$ represents the slack variables for the soft-margin formulation and $\alpha_i$ represents the Lagrange multipliers.

The SVM finds a maximum margin separating hyperplane for the binary classification dataset. This tends to be biased in cases when the dataset is unbalanced towards the class which has more number of samples. This is undesirable for better generalization on such unbalanced datasets. In order to address this situation, the TWSVM evolved as a viable alternative. 

The TWSVM determines two non-parallel separating hyperplanes, each of which is as close as possible to points of one class while being at a distance of at least unity from points of the other class. Let these hyperplanes be denoted by the coefficient vectors $w_{(+1)}, w_{(-1)}$ and bias terms $b_{(+1)}, b_{(-1)}$ corresponding to classes $(+1)$ and $(-1)$ respectively. The TWSVM formulation thus involves solving the set of QPPs given by Eqns. (\ref{twsvm_primal_1})-(\ref{twsvm_primal_2b}).

\begin{eqnarray}
\min_{w_{(+1)},b_{(+1)},q} \frac{1}{2}\|Aw_{(+1)}+e_{+1}b_{(+1)}\|^{2} + C_{(+1)} e_{-1}^{T}q , \label{twsvm_primal_1}\\
s.t. -(Bw_{(+1)}+e_{-1} b_{(+1)}) + q \geq e_{-1} \label{twsvm_primal_1a}\\
 q \geq 0 \label{twsvm_primal_1b}\\
\min_{w_{(-1)},b_{(-1)},q} \frac{1}{2}\|Bw_{(-1)}+e_{-1}b_{(-1)}\|^{2} + C_{(-1)} e_{+1}^{T}q , \label{twsvm_primal_2}\\
s.t. -(Aw_{(-1)}+e_{+1}b_{(-1)}) + q \geq e_{+1} \label{twsvm_primal_2a}\\
 q \geq 0 \label{twsvm_primal_2b}
\end{eqnarray}

Here, data belonging to classes $(+1)$ and $(-1)$ are represented by matrices $A_{[N_A \times M]}$ and $B_{[N_B \times M]}$ respectively, where $N_A$ and $N_B$ are the number of points in each class such that $N_A + N_B = N$. $e_{(+1)}$ and $e_{(-1)}$ are vectors of ones of appropriate dimensions, $q$ represents the error variable associated with the data point and $C_{(+1)}, C_{(-1)} (>0)$ are hyperparameters. It may be noted here that each of the optimization problems solved for the TWSVM has constraints of only the points belonging to one class, as opposed to the conventional SVM which has points of both classes in the constraints. This makes the TWSVM faster than the conventional SVM.

For a test point $x$, the prediction $y$ is computed by a distance measure based on these two hyperplanes. Essentially, the closer the test point is to one of these hyperplanes, the corresponding class is assigned to it. This is summarized by Eqn. (\ref{eqn:twsvmpred}).

\begin{gather}
   y = \left\{
     \begin{array}{lr}
       +1 & : w_{(+1)}^T x+b_{(+1)} \leq w_{(-1)}^T x+b_{(-1)}\\       
       -1 & : otherwise
     \end{array}
   \right.
   \label{eqn:twsvmpred}
\end{gather} 

%

In practice, it is more efficient to solve the dual formulation for the TWSVM, which is given by the Eqns. (\ref{twsvm_dual_1})-(\ref{twsvm_dual_2}).

\begin{eqnarray}
\max_{\alpha} && e_2^T \alpha - \frac{1}{2} \alpha^T G {(H^T H)}^{-1} G^T \alpha \label{twsvm_dual_1}\\
s.t. && 0 \leq  \alpha \leq c_{1} \nonumber \\
\max_{\beta} && e_1^T \beta - \frac{1}{2} \beta^T P {(Q^T Q)}^{-1} P^T \beta \label{twsvm_dual_2}\\
s.t. && 0 \leq  \beta \leq c_{2} \nonumber
\end{eqnarray}

Here, $H = P = [A, e_{(+1)}]$, $G = Q = [B, e_{(-1)}], u=[w_{(+1)} \; b_{(+1)}]^T$, $v=[w_{(-1)} \; b_{(-1)}]^T$. As in the case of SVMs, the kernel formulation for the TWSVM is also possible, where the input dataset $X$ is mapped to another feature space (often infinite dimensional) to introduce linear separability in the dataset samples in the mapped feature space. This is typically accomplished using a mapping function denoted by $\phi(\cdot)$. The separating hyperplanes are then determined in this mapped space, which is often called as the kernel space.

However, it can be seen from the dual formulation of the TWSVM in Eqns. \ref{twsvm_dual_1}-\ref{twsvm_dual_2} that there is a requirement for matrix inversion. This makes it unfeasible for large datasets, where such computations require large memory and are often intractable. Hence, we look for an approach for using the TWSVM within a neural network framework, that would not require such computations thereby making it scalable for large datasets.

\section{The Twin Neural Network Formulation \label{sec:twnn}}

We consider a three-layer neural network to motivate the formulation of the Twin Neural Network, as shown in Fig. \ref{fig:twin_nn}, whose structure is described as follows. The input layer takes the training samples $x^{(i)}, i=1,2,...,N$ and transforms it to a space $\phi(\cdot)$ by the neurons of the hidden layer.  The final or output layer of this network learns a classifier in the feature space denoted by $\phi(\cdot)$, and the classifier hyperplane coefficients (weight vector $w$ and bias $b$ are used to arrive at the prediction for a test sample.

For the case of an unbalanced dataset, we propose the formulation of a Twin Neural Network, where two such networks are trained, whose error functions are denoted by $E_{(+1)}$ and $E_{(-1)}$. These are defined as shown in Eqns. (\ref{twnn1})-(\ref{twnn2}).

\begin{gather}
E_{(+1)} = \frac{1}{2 \times N_B} \sum_{i=1}^{N_B} (t_{(i)} - y_{(i)})^2  \nonumber \\
+ \frac{C_{(+1)}}{2 \times N_A} \sum_{i=1}^{N_A} ( w_{(+1)}^T \phi(x^{(i)}_{(+1)}) + b_{(+1)})^2 \label{twnn1}\\
E_{(-1)} = \frac{1}{2 \times N_A} \sum_{i=1}^{N_A} (t_{(i)} - y_{(i)})^2  \nonumber \\
+ \frac{C_{(-1)}}{2 \times N_B} \sum_{i=1}^{N_B} ( w_{(-1)}^T \phi(x^{(i)}_{(-1)}) + b_{(-1)})^2 \label{twnn2}
\end{gather}

To minimize the error, we set the corresponding derivatives to zero to obtain update rules for the weight vector $w$ and bias $b$. The derivatives w.r.t. $w_{(+1)}$ and $w_{(-1)}$ are shown in Eqns. (\ref{twnn_w1})-(\ref{twnn_w2}), which correspond to the weight update rules for our Twin Neural Network.

\begin{gather}
\frac{\partial E_{(+1)}}{\partial w_{(+1)}} = \frac{1}{N_B} \sum_{i=1}^{N_B} (t_{(i)} - y_{(i)})  (1 - y_{(i)}^2)  \phi(x^{(i)}_{(-1)}) \nonumber  \\
+  \frac{C_{(+1)}}{N_A} \sum_{i=1}^{N_A} (w_{(+1)}^T \phi(x^{(i)}_{(+1)}) + b_{(+1)})  \phi(x^{(i)}_{(+1)}) \label{twnn_w1} \\
\frac{\partial E_{(-1)}}{\partial w_{(-1)}} = \frac{1}{N_A} \sum_{i=1}^{N_A} (t_{(i)} - y_{(i)})  (1 - y_{(i)}^2)  \phi(x^{(i)}_{(-1)}) \nonumber  \\
+  \frac{C_{(-1)}}{N_B} \sum_{i=1}^{N_B} (w_{(-1)}^T \phi(x^{(i)}_{(+1)}) + b_{(-1)})  \phi(x^{(i)}_{(-1)}) \label{twnn_w2}
\end{gather}

Derivatives w.r.t $b_{(+1)}$ and $b_{(-1)}$ are shown in Eqns. (\ref{twnn_b1})-(\ref{twnn_b2}), which are used as the update rule for the bias terms in our Twin Neural Network.

\begin{gather}
\frac{\partial E_{(+1)}}{\partial b_{(+1)}} = \frac{1}{N_B} \sum_{i=1}^{N_B} (t_{(i)} - y_{(i)})  (1 - y_{(i)}^2) \nonumber \\
+ \frac{C_{(+1)}}{N_A} \sum_{i=1}^{N_A}(w_{(+1)}^T \phi(x^{(i)}_{(+1)}) + b_{(+1)}) \label{twnn_b1} \\
\frac{\partial E_{(-1)}}{\partial b_{(-1)}} = \frac{1}{N_A} \sum_{i=1}^{N_A} (t_{(i)} - y_{(i)})  (1 - y_{(i)}^2) \nonumber \\
+ \frac{C_{(-1)}}{N_B} \sum_{i=1}^{N_B}(w_{(-1)}^T \phi(x^{(i)}_{(-1)}) + b_{(-1)}) \label{twnn_b2}
\end{gather}

Based on these, the weights and bias of the hyperplane are updated across the iterations until these hyperplane parameters converge. The prediction on a test point $x$ is obtained as follows. First, the point is mapped to the space $\phi(\cdot)$ by the hidden layer. Then for the output layer, the label $y$ is predicted using Eqn. \ref{eqn:twnnpred}.

\begin{gather}
   y = \left\{
     \begin{array}{lr}
       +1 & : \frac{w_{(+1)}^T \phi(x)+b_{(+1)}}{||w_{(+1)}||} \leq \frac{w_{(-1)}^T \phi(x)+b_{(-1)}}{||w_{(-1)}||}\\
       -1 & : otherwise
     \end{array}
   \right.
   \label{eqn:twnnpred}
\end{gather}

\section{Twin Neural Network for Multi-class Datasets \label{sec:twmc}}

\begin{figure}[hbtp]
\centering
\includegraphics[scale=0.8]{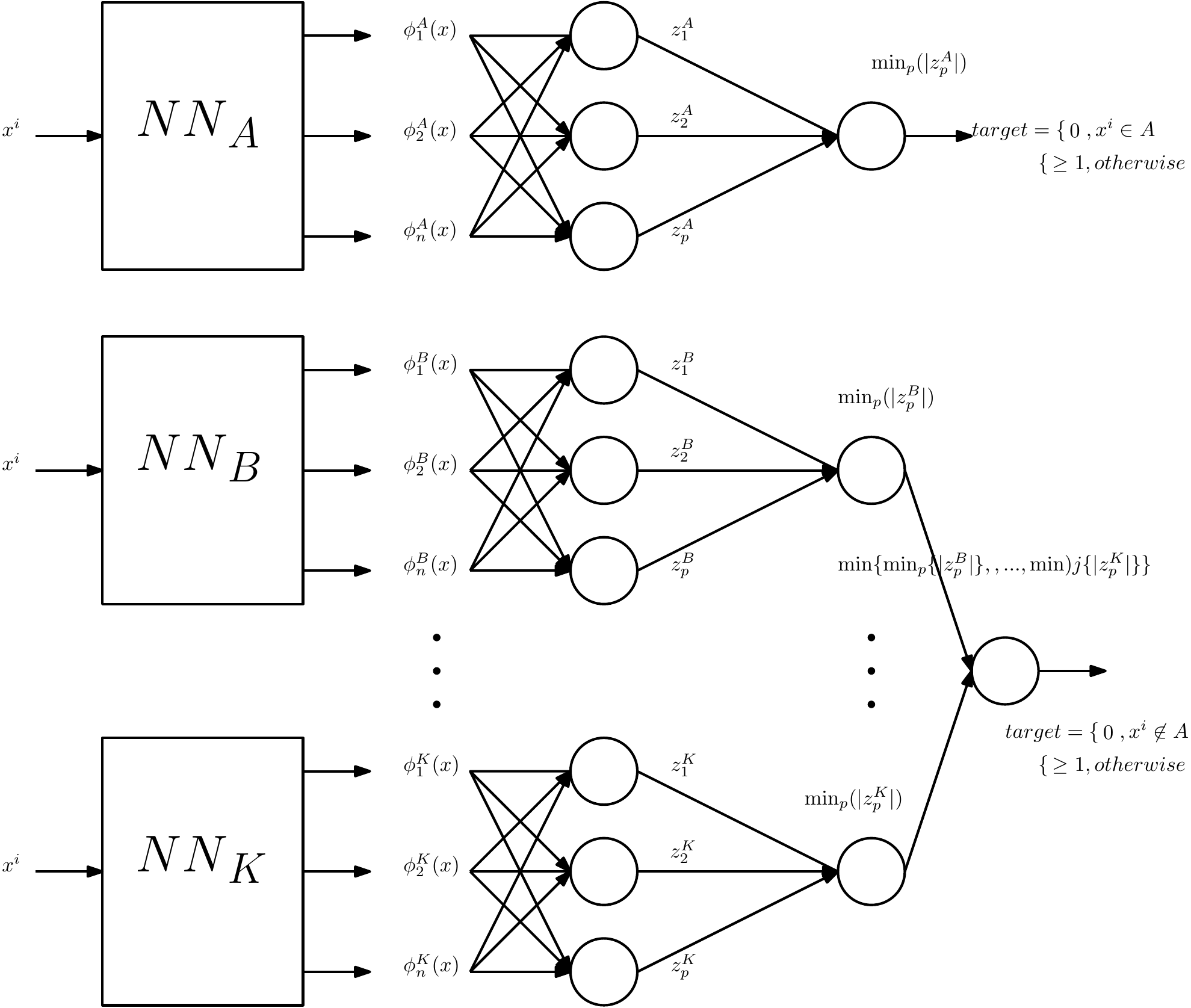}
\caption{Twin NN for multiclass datasets}
\label{fig:twin_nn_mc}
\end{figure}

The Twin Neural Network approach can also be extended to multi-class datasets where the target label $y^i \in \{A,B,..,K\}$ can belong to one among $K$ classes for the various samples. The Twin NN for multi-class datasets is realized by training $K$ different networks, each of which uses a cost function which allows it to correctly classify samples of that class. The architecture is summarily illustrated in Fig. \ref{fig:twin_nn_mc}. Here, each class is associated with multiple hyperplanes, and consequently, multiple classifier neurons, e.g. class A has neurons with outputs labelled as $z^A_1$, $z^A_2$, ..., $z^A_p$. While the figure shows an identical number of output neurons for each class, it is possible to have a different number for each class. Each output neuron for a given class is associated with one hyperplane that passes through a number of points of that class, e.g. it may be imagined as passing through a cluster of samples. The Twin NN allows for multiple planes associated with each class; each of these hyperplanes passes through a number of points of the class. Consider a sample of Class A. The Twin NN requires that for a given sample of class A, the hyperplane closest to that sample is at a small distance from the sample (ideally, a distance of zero). At the same time, amongst all other planes associated with other classes (not A), it requires that the closest plane is at a distance of at least one from the sample of class A. The principle is easily extended to samples of any given class.

As shown in the figure, let ${NN}_A, {NN}_B, ..., {NN}_K$ denote the sub-networks corresponding to each of the classes A, B, .., $K$; the outputs of these sub-networks constitute the features used by the subsequent classifier neurons for the respective class. For example, the features for Class A are denoted by $\{\phi^A_1(x),\phi^A_2(x),...,\phi^A_n(x)\}$. It may be noted that the number of features is denoted by $n$ identically for all classes, but this has been done for lucidity and not as a restriction. The outputs of the classifier neurons for class A are denoted as $z^A_1$, $z^A_2$, ..., $z^A_p$. Note that these classifier neurons use an activation function that passes through the origin, such as the $tanh()$ function. The smallest output of this set is related to the distance of the closest hyperplane (from the class A group of classifier neurons) to the input sample. The $min()$ operator after the classifier neurons computes this function, and the target during training for this is 0. In other words, we would like the closes of all hyperplanes associated with class A to pass through a sample of class A. Similarly, we would like the closest of all hyperplanes associated with all classes other than A to be at a distance of 1 from the sample. Note that this distance is obviously computed in the feature space defined earlier.

\section{Experiments and Discussion \label{sec:expts}}

\subsection{Results on UCI datasets}

The Twin NN was tested on 20 benchmark datasets drawn from the UCI machine learning repository \cite{asuncion2007uci}, which included two class and multi-class datasets. Input features were scaled to lie between -1 and 1. Target set values were kept at +1 and -1 for the respective classes. For multi-class problems, a one versus rest approach \cite[pp. 182, 338]{bishop2006pattern} was used, with as many output layer neurons as the number of classes. A Twin NN with one hidden layer was employed for obtaining these results. The number of neurons in the hidden layer and the hyperparameters were optimized by using a grid search. The K-Nearest Neighbor (KNN) imputation method was used for handling missing attribute values, as it is robust to bias between classes in the dataset. Accuracies were obtained using a standard 5-fold cross validation methodology. This process was repeated 10 times to remove the effect of randomization. The accuracies were compared with the standard SVM  \cite{chang2011libsvm}, TWSVM and Regularized Feed-Forward Neural Networks (RFNN) \cite{bebis1994feed}. The results are shown in Table \ref{tab:res_uci_acc}, which clearly indicates the superior performance of the Twin NN compared to the SVM, TWSVM and and RFNN for 15 of the 20 datasets.

\begin{table*}[htbp]
  \centering
  \caption{Results on UCI datasets for the Twin NN.}
  \scalebox{0.7}{
    \begin{tabular}{|c|c|c|c|c|c|c|c|}
    \hline
    \textbf{S. No.} & \textbf{Dataset} & \textbf{Lin SVM} & \textbf{Ker SVM} & \textbf{RFNN} & \textbf{Twin-NN} & \textbf{Lin TWSVM} & \textbf{Ker TWSVM} \\
    \hline
    1     & Pimaindians (768x4) & 76.5  $\pm$ 2.99 & 77.33  $\pm$ 3.15 & 76.11  $\pm$ 3.60 & \textbf{78.19  $\pm$ 2.73} & 72.99 $\pm$ 6.00 & 75.91 $\pm$ 6.05 \\
    2     & Heartstat (270x13) & 83.33 $\pm$ 4.71 & 84.81  $\pm$ 3.56 & 81.01  $\pm$ 4.82 & \textbf{84.81  $\pm$ 2.74} & 83.11 $\pm$ 5.86 & 82.49 $\pm$ 3.42 \\
    3     & Haberman (306x3) & 72.22 $\pm$ 1.17 & 72.32  $\pm$ 1.18 & 73.11  $\pm$ 2.71 & \textbf{76.11  $\pm$ 4.54} & 73.53  $\pm$ 0.53 & 73.53  $\pm$ 0.53 \\
    4     & Hepatitis (155x19) & 80.00  $\pm$ 6.04 & 83.96  $\pm$ 4.05 & 81.11  $\pm$ 6.29 & \textbf{86.50  $\pm$ 5.98} & 79.11  $\pm$ 4.22 & 82.87  $\pm$ 1.71 \\
    5     & Ionosphere (351x34) & 87.82  $\pm$ 2.11 & \textbf{95.43  $\pm$ 2.35} & 86.21  $\pm$ 4.28 & 94.01  $\pm$ 1.87 & 85.55  $\pm$ 2.93 & 88.92  $\pm$ 1.52 \\
    6     & Transfusion (748x4) & 76.20  $\pm$ 0.27 & 76.60  $\pm$ 0.42 & 76.01  $\pm$ 1.57 & \textbf{78.07  $\pm$ 1.24} & 76.2  $\pm$ 0.20 & 76.60  $\pm$ 0.42 \\
    7     & ECG   (132x12) & 84.90  $\pm$ 5.81 & 87.20  $\pm$ 8.48 & 86.25  $\pm$ 6.64 & \textbf{91.73  $\pm$ 4.75} & 84.32  $\pm$ 3.18 & 84.88  $\pm$ 2.19 \\
    8     & Voting (435x16) & 93.69  $\pm$ 0.96 & \textbf{96.56  $\pm$ 1.13} & 94.47  $\pm$ 1.90 & 96.1  $\pm$ 1.5 & 93.11  $\pm$ 1.41 & 95.32  $\pm$ 1.11 \\
    9     & Fertility (100x9) & 85.03  $\pm$ 6.03 & \textbf{88.03  $\pm$ 2.46} & 87.91  $\pm$ 6.51 & \textbf{88.03  $\pm$ 2.46} & 65.95  $\pm$ 3.97 & 88.03  $\pm$ 2.46 \\
    10    & Australian (690x14) & 85.50 $\pm$ 4.04 & 86.51  $\pm$ 3.96 & 85.24  $\pm$ 3.52 & \textbf{87.97  $\pm$ 3.89} & 85.71 $\pm$ 4.11 & 85.67 $\pm$ 1.02 \\
    11    & CRX   (690x15) & 69.56  $\pm$ 0 & 69.56   $\pm$ 0 & 68.14  $\pm$ 0.94 & \textbf{70.5  $\pm$ 1.62} & 65.56  $\pm$ 0.34 & 69.57   $\pm$ 0 \\
    12    & Mamm-masses (961x5) & 78.87  $\pm$ 2.14 & \textbf{83.25  $\pm$ 3.77} & 77.96  $\pm$ 2.00 & 80.75  $\pm$ 2.37 & 78.51  $\pm$ 1.10 & 80.11  $\pm$ 1.23 \\
    13    & German (1000x20) & 74.1  $\pm$ 2.77 & 75.20  $\pm$ 2.58 & 75.8  $\pm$ 2.88 & \textbf{76.3  $\pm$ 1.35} & 71.99 $\pm$ 5.11 & 72.87 $\pm$ 4.71 \\
    14    & PLRX  (182x12) & 71.44  $\pm$ 1.06 & \textbf{72.52  $\pm$ 0.44} & 71.05  $\pm$ 3.54 & 72.01  $\pm$ 1.94 & 72.08  $\pm$ 6.7 & 71.44 $\pm$ 1.06 \\
    15    & SONAR (208x60) & 76.02  $\pm$ 6.70 & 87.02  $\pm$ 6.47 & 86.62 $\pm$ 6.90 & \textbf{88.53  $\pm$ 5.27} & 76.11  $\pm$ 3.8 & 79.35 $\pm$ 7.11 \\
    16    & Housevotes (436x16) & 95.88 $\pm$ 1.90 & 96.56  $\pm$ 1.13 & 95.56 $\pm$ 1.56 & \textbf{97.02 $\pm$ 1.00} & 94.61  $\pm$ 1.21 & 96.32  $\pm$ 2.71 \\
    17    & Balance (576x4) & 94.61  $\pm$ 1.68 & \textbf{99.82  $\pm$ 0.38} & 97.39  $\pm$ 2.39 & 97.70  $\pm$ 1.93 & 94.99  $\pm$ 1.7 & 97.11  $\pm$ 2.42 \\
    18    & Blogger (100x5) & 70.93  $\pm$ 12.4 & 85.82  $\pm$ 8.48 & 79.50  $\pm$ 9.35 & \textbf{86.01  $\pm$ 4.22} & 72.11  $\pm$ 1.03 & 80.87  $\pm$ 1.11 \\
    19    & IPLD  (583x10) & 71.35  $\pm$ 0.39 & 71.35  $\pm$ 0.39 & 71.05  $\pm$ 4.20 & \textbf{73.11  $\pm$ 1.64} & 69.97  $\pm$ 1.16 & 71.35  $\pm$ 0.48 \\
    20    & Heart Spectf (267x44) & 78.89  $\pm$ 1.02 & 79.16  $\pm$ 1.23 & 79.03  $\pm$ 1.17 & \textbf{83.34  $\pm$ 3.4} & 78.41  $\pm$ 2.17 & 79.51  $\pm$ 1.70 \\
    \hline
    \end{tabular}%
    }
  \label{tab:res_uci_acc}%
\end{table*}%

In addition, we also present a comparative analysis of the performance of the Twin NN on UCI benchmark datasets w.r.t. other  approaches in terms of p-values determined using Wilcoxon's signed ranks test \cite{wilcoxon1945individual}. The values for the Wilcoxon signed ranks test. The Wilcoxon Signed-Ranks Test is a measure of the extent of statistical deviation in the results obtained by using an approach. A p-value less than $0.05$ indicates that the results have a significant statistical difference with the results obtained using the reference approach, whereas p-values greater than $0.05$ indicate non-significant statistical difference. The p-values for the approaches considered are shown in Table \ref{tab:res_uci_wilcoxon}, which clearly indicates that the Twin NN works better than the reference approaches.

\begin{table}[h]
  \centering
  \caption{Wilcoxon signed ranks test for the Twin NN}
    \begin{tabular}{|c||c|c|}
    \hline
    \textbf{S.No} & \textbf{Algorithm} & \textbf{p value} \\
    \hline
    1     & Lin SVM & 8.18E-05 \\
    2     & Ker SVM & 3.80E-02 \\
    3     & RFNN  & 8.84E-05 \\
    4     & Lin TWSVM & 1.01E-04 \\
    5     & Ker TWSVM & 1.31E-04 \\
    \hline
    \end{tabular}%
  \label{tab:res_uci_wilcoxon}%
\end{table}%

To verify that the results on the UCI datasets are independent of the randomization due to the data distribution across folds, we also performed the Friedman's test \cite{friedman1937use}. The p-value is $3.77\times 10^{-14}$ ($<0.05$), implying that the superior results of the Twin NN are not due to chance, and are clearly a consequence of the approach used rather than the datasets.

\subsection{Results on highly unbalanced datasets}

\begin{figure}[hbtp]
\centering
\includegraphics[scale=0.8]{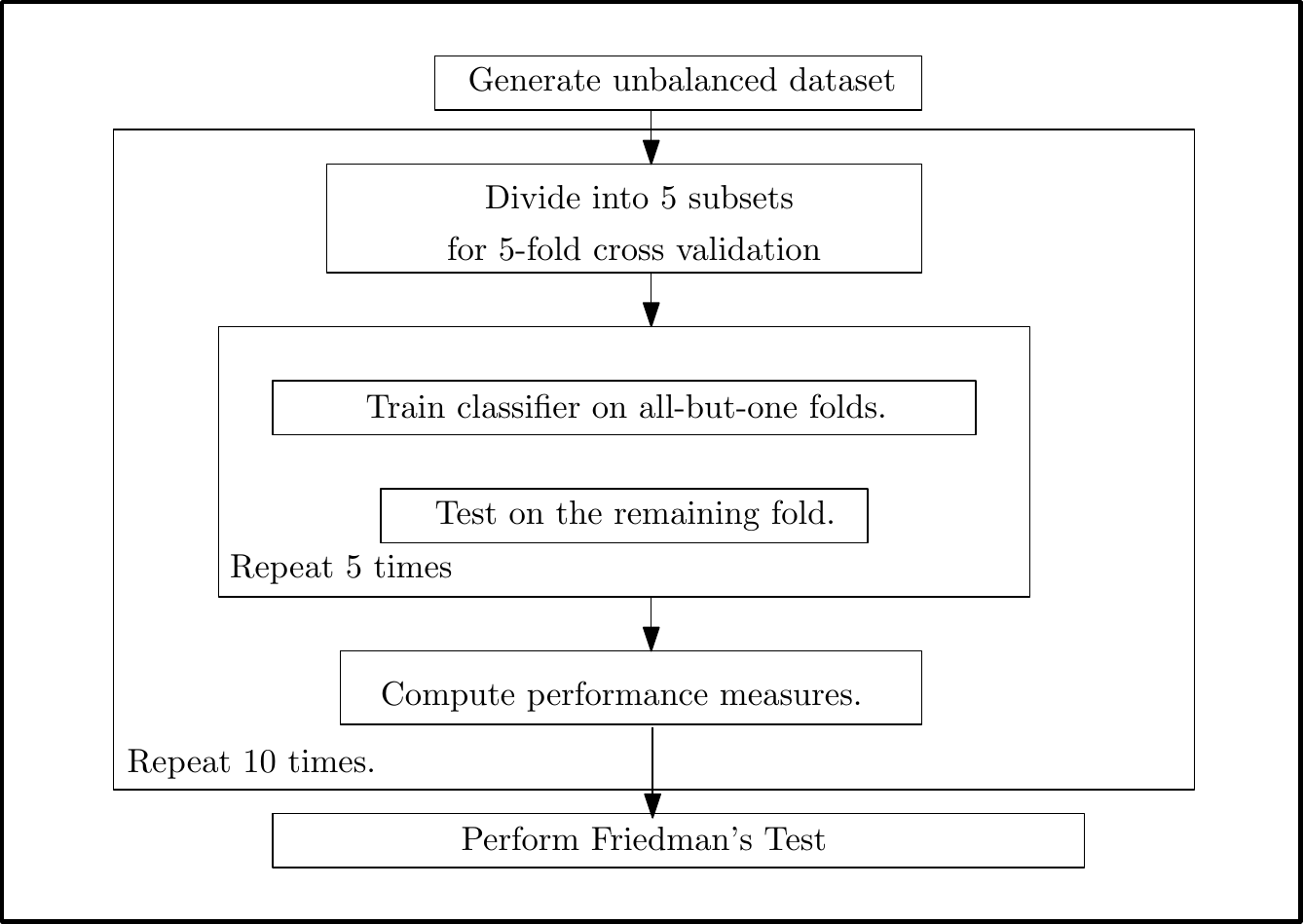}
\caption{Flowchart for computing accuracy of unbalanced dataset.}
\label{fig:unbal_expt_proc}
\end{figure}

The key benefit obtained by using the Twin NN is better generalization for unbalanced datasets. To establish this, we evaluate the Twin NN on several unbalanced datasets \cite{kubat1998machine, woods1993comparative, ding2011diversified}, which are summarized in Table \ref{tab:res_imbal_datasets}. It may be noted here that the class imbalance has been introduced in these datasets by considering the multi-class datasets as separate binary datasets using one-v/s-rest approach. Thus for a dataset having $N$ classes and $M$ samples per class, we can in principle generate $N$ datasets, each of which has a class ratio of $M:(N-1)\cdot M$. Each of these possible datasets has been denoted as ``Gen$N$'' suffixed to the dataset name in Table \ref{tab:res_imbal_datasets}, where $N$ represents the corresponding class w.r.t. which imbalance has been induced in the dataset. The procedure to compute the accuracies is illustrated in Fig. \ref{fig:unbal_expt_proc}.

\begin{table*}[htbp]
  \centering
  \caption{Description of large unbalanced datasets, attributes are numeric (N) or categorical (C)}
  \scalebox{0.7}{
    \begin{tabular}{|c|c|c|c|c|c|c|}
    \hline
    S. No. & Dataset Generated & Source & Area  & Sample Ratio & No of Samples & Attributes \\
    \hline
    1     & Abalone\_Gen1 & UCI, link & Life  & 1:9 & 391 : 3,786 & 1N,7C \\
    2     & Letter\_Gen & UCI   & Computer & 1:26 & 734 :19,266 & 16 \\
    3     & Yeast\_Gen & UCI   & Life  & 1:28  & 51 : 1,433 & 8 \\
    4     & Abalone\_Gen2 & UCI   & Life  & 1:129 & 32 : 4,145 & 8 \\
    5     & Coil\_2000 & UCI KDD & Business & 1:15 & 586 : 9,236 & 85 \\
    6     & Car\_Eval\_Gen1 & UCI   & Business & 1:25 & 65 : 1,663 & 6N, 21 C \\
    7     & Wine\_Quality\_Gen & UCI   & Chemistry & 1:26 & 181 : 4,715 & 11C \\
    8     & Forest\_CovType\_Gen & UCI KDD & Nature & 1:210 & 2747 : 578,265 & 44N,10C \\
    9     & Ozone\_Level & UCI   & Environment & 1:33 & 73 : 2,463 & 72C \\
    10    & Pen Digits\_Gen & UCI   & Computer & 1:9 & 1,055 : 9,937 & 16C \\
    11    & Spectrometer\_Gen & UCI   & Physics & 1:10 & 45 : 486 & 93C \\
    12    & Statlog\_Gen & UCI   & Nature & 1:9 & 626 : 5,809 & 36C \\
    13    & Libras\_Gen & UCI   & Physics & 1:14  & 24 :336 & 90C \\
    14    & Optical Digits\_Gen & UCI   & Computer & 1:9 & 554 : 5,066 & 64C \\
    15    & Ecoli\_Gen & UCI   & Life  & 1:8 & 35 : 301 & 7C \\
    16    & Car Evaluation\_Gen2 & UCI   & Business & 1:11 & 134 : 1,594 & 6N \\
    17    & US\_Crime\_Gen & UCI   & Economics & 1:12 & 150 : 1,844 & 122C \\
    18    & Protein\_homology & KDD CUP 2004 & Biology & 1:111 & 1,296 : 144,455 & 74C \\
    19    & Scene\_Gen & LibSVM Data & Nature & 1:12 & 177 : 2,230 & 294C \\
    20    & Solar\_Flare\_Gen & UCI   & Nature & 1:19 & 68 : 1,321 & 10N \\
    \hline
    \end{tabular}%
    }
  \label{tab:res_imbal_datasets}%
\end{table*}%

\begin{figure}[H]
\centering
\includegraphics[scale=0.7]{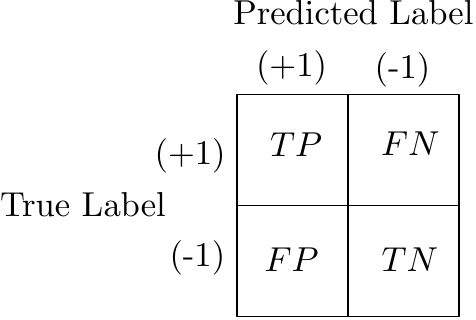}
\caption{Confusion Matrix}
\label{fig:confusion_matrix}
\end{figure}

For evaluating the performance of the Twin NN on unbalanced datasets, measures other than the classification accuracy are often used in practice. These provide a better understanding of the performance of the learning algorithm taking into consideration the skewness in the dataset. To define these measures, we use four basic quantities, True Positives ($TP$), True Negatives ($TN$), False Positives ($FP$) and the False Negatives ($FN$). $TP$ is the number of samples whose class label and predicted label are ``true'' and $TN$ is the number of samples whose class and predicted labels are ``false''. Correspondingly, we use $FP$ for the samples whose class label is ``false'' and predicted label is ``true'', and vice-versa for $FN$. These are schematically illustrated in the confusion matrix as shown in Fig. \ref{fig:confusion_matrix}.

Based on these, the basic measures are defined as given by Eqns. (\ref{acc})-(\ref{npv}). Also,  $PC=TP+FN$, $NC=FP+TN$, $PR=TP+FP$ and $NR=FN+TN$.

\begin{gather}
\text{Accuracy (Acc):=}\; \frac{TP+TN}{TP+TN+FP+FN} \label{acc}
\end{gather}
\begin{gather}
\text{True Positive Rate (TPR):=}\; \frac{TP}{TP+FN} = \frac{TP}{PC}\label{tpr}\\
\text{True Negative Rate (TNR):=}\; \frac{TN}{TN+FP} = \frac{TN}{NC}\label{tnr}\\
\text{Positive Prediction Value (PPV):=}\; \frac{TP}{TP+FP} = \frac{TP}{PR}\label{ppv}\\
\text{Negative Prediction Value (NPV):=}\; \frac{TN}{TN+FN} = \frac{TN}{NR}\label{npv}
\end{gather}

The performance measure in terms of geometric mean between accuracy of the two classes is called the G-means. This is computed as the square-root of the product of the true positive and true negative rates, or as $\sqrt{TPR*TNR}$. The G-means for the unbalanced datasets is shown in Table \ref{tab:res_imbal_gmeans}. One can observe that the Twin NN has higher G-means value for all except one dataset when compared to the competing approaches. 

\begin{table*}[htbp]
  \centering
  \caption{G-means of large datasets with high unbalance}
   \scalebox{0.7}{
    \begin{tabular}{|c|c|c|c|c|c|c|c|}
    \hline
    S. No. & Dataset & Twin NN & Lin SVM & Ker SVM & RFNN  & Lin TWSVM & Ker TWSVM  \\
    \hline
    1     & Abalone\_Gen1 & \textbf{0.747 $\pm$  0.03} & 0.045 $\pm$  0.01 & 0.331  $\pm$  0.14 & 0.04 $\pm$  0.015 & \multicolumn{1}{|c|}{0.127 $\pm$  0.02} & \multicolumn{1}{|c|}{0.358  $\pm$  0.11} \\
    2     & Letter\_Gen & 0.95  $\pm$  0.06 & 0.83 $\pm$  0.05 & \textbf{0.98  $\pm$  0.09} & 0.88 $\pm$  0.01 &  -  & -\\
    3     & Yeast\_Gen & \textbf{0.84  $\pm$  0.04} & 0.2106  $\pm$  0.09 & 0.411  $\pm$  0.10 & 0.47  $\pm$  0.12 & \multicolumn{1}{|c|}{0.177  $\pm$  0.04} & \multicolumn{1}{|c|}{0.31  $\pm$  0.06} \\
    4     & Abalone\_Gen2 & \textbf{0.36  $\pm$  0.11} & 0 $\pm$  0  & 0  $\pm$  0 & 0 $\pm$  0  & \multicolumn{1}{|c|}{0.20 $\pm$  0.04} & \multicolumn{1}{|c|}{0.205 $\pm$  0.01} \\
    5     & Coil\_2000 & \textbf{0.66  $\pm$  0.01} & 0 $\pm$  0  & 0.031  $\pm$  0.07 & 0.055  $\pm$  0.05 &    - & - \\
    6     & Car\_Eval\_Gen1 & \textbf{1  $\pm$  0.00} & 0.934  $\pm$  0.06 & 0.999  $\pm$  0.001 & 0.999  $\pm$  0.001 & \multicolumn{1}{|c|}{0.935  $\pm$  0.04} & \multicolumn{1}{|c|}{0.955  $\pm$  0.08} \\
    7     & Wine\_Quality\_Gen & \textbf{0.735  $\pm$  0.07} & 0  $\pm$  0 & 0.35  $\pm$  0.05 & 0.245  $\pm$  0.13 & \multicolumn{1}{|c|}{0.25  $\pm$  0.04} & \multicolumn{1}{|c|}{0.29 $\pm$  0.01} \\
    8     & Ozone\_Level & \textbf{0.84 $\pm$  0.008} & 0 $\pm$  0  & 0.283 $\pm$  0.04 & 0.53 $\pm$  0.06 & \multicolumn{1}{|c|}{0.187 $\pm$  0.024} & \multicolumn{1}{|c|}{0.479 $\pm$  0.10} \\
    9     & Pen Digits\_Gen & \textbf{0.9927 $\pm$  0.02} & 0.847 $\pm$  0.01 & 0.983 $\pm$  0.06 & 0.77 $\pm$  0.022 &   -    & - \\
    10    & Spectrometer\_Gen & \textbf{0.956 $\pm$  0.05} & 0.902 $\pm$  0.05 & 0.949 $\pm$  0.05 & 0.938 $\pm$  0.04 & \multicolumn{1}{|c|}{0.91 $\pm$  0.07} & \multicolumn{1}{|c|}{0.950 $\pm$  0.09} \\
    11    & Statlog\_Gen & \textbf{0.890 $\pm$  0.011} & 0.137 $\pm$  0.04 & 0.745 $\pm$  0.023 & 0.705 $\pm$  0.023 & \multicolumn{1}{|c|}{0.22 $\pm$  0.01} & \multicolumn{1}{|c|}{0.68 $\pm$  0.01} \\
    12    & Libras\_Gen & \textbf{0.935 $\pm$  0.05} & 0.866 $\pm$  0.09 & 0.935 $\pm$  0.05 & 0.89 $\pm$  0.07 & \multicolumn{1}{|c|}{0.821 $\pm$  0.05} & \multicolumn{1}{|c|}{0.947 $\pm$  0.0} \\
    13    & Optical Digits\_Gen & \textbf{0.991 $\pm$  0.007} & 0.888 $\pm$  0.02 & 0.98 $\pm$  0.007 & 0.964 $\pm$  0.017 & \multicolumn{1}{|c|}{0.89 $\pm$  0.01} & \multicolumn{1}{|c|}{0.952 $\pm$  0.002} \\
    14    & Ecoli\_Gen & \textbf{0.900 $\pm$  0.06} & 0.69 $\pm$  0.11 & 0.78 $\pm$  0.12 & 0.83 $\pm$  0.06 & \multicolumn{1}{|c|}{0.697 $\pm$  0.017} & \multicolumn{1}{|c|}{0.825 $\pm$  0.01} \\
    15    & Car Evaluation\_Gen2 & \textbf{0.996 $\pm$  0.002} & 0.949 $\pm$  0.05 & 0.976 $\pm$  0.04 & 0.990 $\pm$  0.011 & \multicolumn{1}{|c|}{0.934 $\pm$  0.07} & \multicolumn{1}{|c|}{0.944 $\pm$  0.007} \\
    16    & US\_Crime\_Gen & \textbf{0.85 $\pm$  0.03} & 0.635 $\pm$  0.03 & 0.63 $\pm$  0.03 & 0.69 $\pm$  0.04 & \multicolumn{1}{|c|}{0.58 $\pm$  0.01} & \multicolumn{1}{|c|}{0.65 $\pm$  0.01} \\
    17    & Protein\_homology & \textbf{0.942 $\pm$  0.006} & 0.85 $\pm$  0.01 & 0.89 $\pm$  0.01 & 0.871 $\pm$  0.006 & -      &  -\\
    18    & Scene\_Gen & \textbf{0.712 $\pm$  0.04} & 0.066 $\pm$  0.09 & 0.321 $\pm$  0.06 & 0.389 $\pm$  0.04 & \multicolumn{1}{|c|}{0.321 $\pm$  0.03} & \multicolumn{1}{|c|}{0.39 $\pm$  0.01} \\
    19    & Solar\_Flare\_Gen & \textbf{0.705 $\pm$  0.03} & 0  $\pm$  0 & 0  $\pm$  0 & 0.29  $\pm$  0.04 & \multicolumn{1}{|c|}{0.35 $\pm$  0.005} & \multicolumn{1}{|c|}{0.393 $\pm$  0.05} \\
    \hline
    \end{tabular}%
    }
  \label{tab:res_imbal_gmeans}%
\end{table*}%

The harmonic mean between the positive prediction value and true positive rate is called the F-measure, which is computed as given by Eqn. (\ref{fmeasure}).
\begin{gather}
\text{F-measure}\;:=\;\frac{2}{\frac{1}{TPR}+\frac{1}{PPV}} \label{fmeasure}
\end{gather}

The f-measure for the unbalanced datasets is shown in Table \ref{tab:res_imbal_fmeasure}. From the results, it is evident that the Twin NN has higher f-measure values for all except one dataset, which demonstrates its superior performance.
\begin{table*}[htbp]
  \centering
  \caption{F-measure for large datasets with high unbalance}
  \scalebox{0.7}{
    \begin{tabular}{|c|c|c|c|c|c|c|c|}
    \hline
    S. No. & Dataset & Twin NN & Lin SVM & Ker SVM & RFNN  & Lin TWSVM & Ker TWSVM \\
    \hline
    \multicolumn{1}{|c|}{1} & \multicolumn{1}{|c|}{Abalone\_Gen1} & \multicolumn{1}{|c|}{\textbf{0.358  $\pm$  0.05}} & \multicolumn{1}{|c|}{0.018  $\pm$  0.005} & \multicolumn{1}{|c|}{0.065  $\pm$  0.01} & \multicolumn{1}{|c|}{0.0088  $\pm$  0.007} & \multicolumn{1}{|c|}{0.171  $\pm$  0.0007} & \multicolumn{1}{|c|}{0.212  $\pm$  0.01} \\
    \multicolumn{1}{|c|}{2} & \multicolumn{1}{|c|}{Letter\_Gen} & \multicolumn{1}{|c|}{0.861  $\pm$  0.02} & \multicolumn{1}{|c|}{0.74  $\pm$  0.08} & \multicolumn{1}{|c|}{\textbf{0.97  $\pm$  0.06}} & \multicolumn{1}{|c|}{0.852  $\pm$  0.02} & -     & - \\
    \multicolumn{1}{|c|}{3} & \multicolumn{1}{|c|}{Yeast\_Gen} & \multicolumn{1}{|c|}{\textbf{0.40  $\pm$  0.08}} & \multicolumn{1}{|c|}{0.092  $\pm$  0.04} & \multicolumn{1}{|c|}{0.278  $\pm$  0.09} & \multicolumn{1}{|c|}{0.33  $\pm$  0.13} & \multicolumn{1}{|c|}{0.066  $\pm$  0.02} & \multicolumn{1}{|c|}{0.19  $\pm$  0.01} \\
    \multicolumn{1}{|c|}{4} & \multicolumn{1}{|c|}{Abalone\_Gen2} & \multicolumn{1}{|c|}{\textbf{0.02  $\pm$  0.008}} & \multicolumn{1}{|c|}{0  $\pm$  0} & \multicolumn{1}{|c|}{0  $\pm$  0} & \multicolumn{1}{|c|}{0  $\pm$  0} & \multicolumn{1}{|c|}{0.015  $\pm$  0.002} & \multicolumn{1}{|c|}{0.0158  $\pm$  0.001} \\
    \multicolumn{1}{|c|}{5} & \multicolumn{1}{|c|}{Coil\_2000} & \multicolumn{1}{|c|}{\textbf{0.19  $\pm$  0.01}} & \multicolumn{1}{|c|}{0  $\pm$  0} & \multicolumn{1}{|c|}{0.021  $\pm$  0.009} & \multicolumn{1}{|c|}{0.015  $\pm$  0.0034} & -     & - \\
    \multicolumn{1}{|c|}{6} & \multicolumn{1}{|c|}{Car\_Eval\_Gen1} & \multicolumn{1}{|c|}{\textbf{1  $\pm$  0.00}} & \multicolumn{1}{|c|}{0.872  $\pm$  0.07} & \multicolumn{1}{|c|}{0.99  $\pm$  0.0005} & \multicolumn{1}{|c|}{0.999  $\pm$  0.0005} & \multicolumn{1}{|c|}{0.892  $\pm$  0.004} & \multicolumn{1}{|c|}{0.937  $\pm$  0.001} \\
    \multicolumn{1}{|c|}{7} & \multicolumn{1}{|c|}{Wine\_Quality\_Gen} & \multicolumn{1}{|c|}{\textbf{0.392 $\pm$  0.01}} & \multicolumn{1}{|c|}{0 $\pm$  0} & \multicolumn{1}{|c|}{0.206 $\pm$  0.05} & \multicolumn{1}{|c|}{0.13 $\pm$  0.08} & \multicolumn{1}{|c|}{0.05 $\pm$  0.004} & \multicolumn{1}{|c|}{0.1 $\pm$  0.001} \\
    \multicolumn{1}{|c|}{8} & \multicolumn{1}{|c|}{Ozone\_Level} & \multicolumn{1}{|c|}{\textbf{0.42  $\pm$  0.05}} & \multicolumn{1}{|c|}{0 $\pm$  0} & \multicolumn{1}{|c|}{0.140  $\pm$  0.03} & \multicolumn{1}{|c|}{0.35  $\pm$  0.08} & \multicolumn{1}{|c|}{0.091  $\pm$  0.001} & \multicolumn{1}{|c|}{0.12  $\pm$  0.01} \\
    \multicolumn{1}{|c|}{9} & \multicolumn{1}{|c|}{Pen Digits\_Gen} & \multicolumn{1}{|c|}{\textbf{0.975  $\pm$  0.01}} & \multicolumn{1}{|c|}{0.975  $\pm$  0.01} & \multicolumn{1}{|c|}{0.975  $\pm$  0.01} & \multicolumn{1}{|c|}{0.743  $\pm$  0.01} & -     & - \\
    \multicolumn{1}{|c|}{10} & \multicolumn{1}{|c|}{Spectrometer\_Gen} & \multicolumn{1}{|c|}{\textbf{0.921  $\pm$  0.08}} & \multicolumn{1}{|c|}{0.869  $\pm$  0.077} & \multicolumn{1}{|c|}{0.915  $\pm$  0.085} & \multicolumn{1}{|c|}{0.899  $\pm$  0.08} & \multicolumn{1}{|c|}{0.872 $\pm$  0.089} & \multicolumn{1}{|c|}{0.918 $\pm$  0.006} \\
    \multicolumn{1}{|c|}{11} & \multicolumn{1}{|c|}{Statlog\_Gen} & \multicolumn{1}{|c|}{\textbf{0.67 $\pm$  0.01}} & \multicolumn{1}{|c|}{0.04 $\pm$  0.02} & \multicolumn{1}{|c|}{0.655 $\pm$  0.02} & \multicolumn{1}{|c|}{0.575 $\pm$  0.03} & \multicolumn{1}{|c|}{0.039 $\pm$  0.005} & \multicolumn{1}{|c|}{0.281 $\pm$  0.01} \\
    \multicolumn{1}{|c|}{12} & \multicolumn{1}{|c|}{Libras\_Gen} & \multicolumn{1}{|c|}{\textbf{0.915 $\pm$  0.04}} & \multicolumn{1}{|c|}{0.83 $\pm$  0.10} & \multicolumn{1}{|c|}{0.90 $\pm$  0.15} & \multicolumn{1}{|c|}{0.88 $\pm$  0.08} & \multicolumn{1}{|c|}{0.81 $\pm$  0.13} & \multicolumn{1}{|c|}{0.124 $\pm$  0.01} \\
    \multicolumn{1}{|c|}{13} & \multicolumn{1}{|c|}{Optical Digits\_Gen} & \multicolumn{1}{|c|}{\textbf{0.984 $\pm$  0.0007}} & \multicolumn{1}{|c|}{0.842 $\pm$  0.018} & \multicolumn{1}{|c|}{0.983 $\pm$  0.008} & \multicolumn{1}{|c|}{0.956 $\pm$  0.018} & \multicolumn{1}{|c|}{0.846 $\pm$  0.019} & \multicolumn{1}{|c|}{0.886 $\pm$  0.030} \\
    \multicolumn{1}{|c|}{14} & \multicolumn{1}{|c|}{Ecoli\_Gen} & \multicolumn{1}{|c|}{\textbf{0.695 $\pm$  0.10}} & \multicolumn{1}{|c|}{0.57 $\pm$  0.16} & \multicolumn{1}{|c|}{0.670 $\pm$  0.15} & \multicolumn{1}{|c|}{0.682 $\pm$  0.11} & \multicolumn{1}{|c|}{0.595 $\pm$  0.19} & \multicolumn{1}{|c|}{0.647 $\pm$  0.15} \\
    \multicolumn{1}{|c|}{15} & \multicolumn{1}{|c|}{Car Evaluation\_Gen2} & \multicolumn{1}{|c|}{\textbf{0.958 $\pm$  0.04}} & \multicolumn{1}{|c|}{0.909 $\pm$  0.07} & \multicolumn{1}{|c|}{0.946 $\pm$  0.06} & \multicolumn{1}{|c|}{0.95 $\pm$  0.04} & \multicolumn{1}{|c|}{0.89 $\pm$  0.09} & \multicolumn{1}{|c|}{0.897 $\pm$  0.02} \\
    \multicolumn{1}{|c|}{16} & \multicolumn{1}{|c|}{US\_Crime\_Gen} & \multicolumn{1}{|c|}{\textbf{0.586 $\pm$  0.04}} & \multicolumn{1}{|c|}{0.50 $\pm$  0.04} & \multicolumn{1}{|c|}{0.497 $\pm$  0.08} & \multicolumn{1}{|c|}{0.55 $\pm$  0.04} & \multicolumn{1}{|c|}{0.45 $\pm$  0.07} & \multicolumn{1}{|c|}{0.56 $\pm$  0.0002} \\
    \multicolumn{1}{|c|}{17} & \multicolumn{1}{|c|}{Protein\_homology} & \multicolumn{1}{|c|}{\textbf{0.868 $\pm$  0.007}} & \multicolumn{1}{|c|}{0.8 $\pm$  0.01} & \multicolumn{1}{|c|}{0.86 $\pm$  0.01} & \multicolumn{1}{|c|}{0.76 $\pm$  0.007} & -     & - \\
    \multicolumn{1}{|c|}{18} & \multicolumn{1}{|c|}{Scene\_Gen} & \multicolumn{1}{|c|}{\textbf{0.337 $\pm$  0.06}} & \multicolumn{1}{|c|}{0.021 $\pm$  0.009} & \multicolumn{1}{|c|}{0.181 $\pm$  0.061} & \multicolumn{1}{|c|}{0.207 $\pm$  0.06} & \multicolumn{1}{|c|}{0.136 $\pm$  0.001} & \multicolumn{1}{|c|}{0.182 $\pm$  0.001} \\
    \multicolumn{1}{|c|}{19} & \multicolumn{1}{|c|}{Solar\_Flare\_Gen} & \multicolumn{1}{|c|}{\textbf{0.30 $\pm$  0.03}} & \multicolumn{1}{|c|}{0  $\pm$  0} & \multicolumn{1}{|c|}{0  $\pm$  0} & \multicolumn{1}{|c|}{0.13  $\pm$  0.02} & \multicolumn{1}{|c|}{0.15 $\pm$  0.01} & \multicolumn{1}{|c|}{0.16 $\pm$  0.005} \\
    \hline
    \end{tabular}%
    }
  \label{tab:res_imbal_fmeasure}%
\end{table*}%

In addition, we also compute the Mathew Correlation Coefficient (MCC) for evaluating the performance of the Twin NN on unbalanced datasets. High MCC values imply that the classifier has high classification accuracy for both classes and also less mis-classification on both classes. The MCC is computed as given by Eqn. \ref{mcc}. The MCC values for the unbalanced datasets is given in Table \ref{tab:imbal_res_mcc}, and the values are higher for the Twin NN for all except two datasets.

\begin{gather}
\text{MCC}\;:=\; \frac{TP*TN - FP*FN}{\sqrt{PC*NC*PR*NR}} \label{mcc}
\end{gather}

\begin{table*}[htbp]
  \centering
  \caption{MCC for large datasets with high unbalance}
  \scalebox{0.7}{
    \begin{tabular}{|c|c|c|c|c|c|c|c|}
    \hline
    S. No. & Dataset Generated & \textbf{Twin NN} & \textbf{Lin SVM} & \textbf{Ker SVM} & \textbf{RFNN} & \textbf{Lin TWSVM} & \textbf{Ker TWSVM} \\
    \hline
    \multicolumn{1}{|c|}{1} & \multicolumn{1}{|c|}{Abalone\_Gen1} & \multicolumn{1}{|c|}{\textbf{0.317 $\pm$  0.05}} & \multicolumn{1}{|c|}{0.019  $\pm$  0.04} & \multicolumn{1}{|c|}{0.05 $\pm$  0.13} & \multicolumn{1}{|c|}{0.043  $\pm$  0.059} & \multicolumn{1}{|c|}{0.08 $\pm$  0.06} & \multicolumn{1}{|c|}{0.125 $\pm$  0.03} \\
    \multicolumn{1}{|c|}{2} & \multicolumn{1}{|c|}{Letter\_Gen} & \multicolumn{1}{|c|}{0.868  $\pm$  0.02} & \multicolumn{1}{|c|}{0.761  $\pm$  0.03} & \multicolumn{1}{|c|}{\textbf{0.97 $\pm$  0.006}} & \multicolumn{1}{|c|}{0.851  $\pm$  0.02} & -     & - \\
    \multicolumn{1}{|c|}{3} & \multicolumn{1}{|c|}{Yeast\_Gen} & \multicolumn{1}{|c|}{\textbf{0.41  $\pm$  0.07}} & \multicolumn{1}{|c|}{0.12   $\pm$  0.16} & \multicolumn{1}{|c|}{0.372  $\pm$  0.068} & \multicolumn{1}{|c|}{0.38   $\pm$  0.10} & \multicolumn{1}{|c|}{0.08  $\pm$  0.07} & \multicolumn{1}{|c|}{0.21  $\pm$  0.08} \\
    \multicolumn{1}{|c|}{4} & \multicolumn{1}{|c|}{Abalone\_Gen2} & \multicolumn{1}{|c|}{\textbf{0.07  $\pm$  0.02}} & \multicolumn{1}{|c|}{0   $\pm$  0} & \multicolumn{1}{|c|}{0  $\pm$  0} & \multicolumn{1}{|c|}{0   $\pm$  0} & \multicolumn{1}{|c|}{0   $\pm$  0} & \multicolumn{1}{|c|}{1   $\pm$  0} \\
    \multicolumn{1}{|c|}{5} & \multicolumn{1}{|c|}{Coil\_2000} & \multicolumn{1}{|c|}{\textbf{0.17  $\pm$  0.01}} & \multicolumn{1}{|c|}{(-)0.0011   $\pm$  0.002} & \multicolumn{1}{|c|}{0.017  $\pm$  0.05} & \multicolumn{1}{|c|}{0.03  $\pm$  0.04} & -     & - \\
    \multicolumn{1}{|c|}{6} & \multicolumn{1}{|c|}{Car\_Eval\_Gen1} & \multicolumn{1}{|c|}{\textbf{1 $\pm$  0.00}} & \multicolumn{1}{|c|}{0.871  $\pm$  0.069} & \multicolumn{1}{|c|}{0.995  $\pm$  0.001} & \multicolumn{1}{|c|}{0.999  $\pm$  0.001} & \multicolumn{1}{|c|}{0.851   $\pm$  0.01} & \multicolumn{1}{|c|}{0.94   $\pm$  0.04} \\
    \multicolumn{1}{|c|}{7} & \multicolumn{1}{|c|}{Wine\_Quality\_Gen} & \multicolumn{1}{|c|}{\textbf{0.41 $\pm$  0.01}} & \multicolumn{1}{|c|}{(-)0.001   $\pm$  0.002} & \multicolumn{1}{|c|}{0.27  $\pm$  0.06} & \multicolumn{1}{|c|}{0.19  $\pm$  0.11} & \multicolumn{1}{|c|}{0  $\pm$  0.05} & \multicolumn{1}{|c|}{0.05  $\pm$  0.02} \\
    \multicolumn{1}{|c|}{8} & \multicolumn{1}{|c|}{Ozone\_Level} & \multicolumn{1}{|c|}{\textbf{0.41  $\pm$  0.05}} & \multicolumn{1}{|c|}{(-)0.0015   $\pm$  0.003} & \multicolumn{1}{|c|}{0.204  $\pm$  0.066} & \multicolumn{1}{|c|}{0.343  $\pm$  0.08} & \multicolumn{1}{|c|}{0.01 $\pm$  0.13} & \multicolumn{1}{|c|}{0.052 $\pm$  0.05} \\
    \multicolumn{1}{|c|}{9} & \multicolumn{1}{|c|}{Pen Digits\_Gen} & \multicolumn{1}{|c|}{\textbf{0.972  $\pm$  0.01}} & \multicolumn{1}{|c|}{0.775  $\pm$  0.02} & \multicolumn{1}{|c|}{0.968  $\pm$  0.01} & \multicolumn{1}{|c|}{0.752  $\pm$  0.018} & -     & - \\
    \multicolumn{1}{|c|}{10} & \multicolumn{1}{|c|}{Spectrometer\_Gen} & \multicolumn{1}{|c|}{\textbf{0.92  $\pm$  0.04}} & \multicolumn{1}{|c|}{0.863  $\pm$  0.08} & \multicolumn{1}{|c|}{0.91  $\pm$  0.08} & \multicolumn{1}{|c|}{0.89  $\pm$  0.08} & \multicolumn{1}{|c|}{0.852  $\pm$  0.09} & \multicolumn{1}{|c|}{0.911  $\pm$  0.05} \\
    \multicolumn{1}{|c|}{11} & \multicolumn{1}{|c|}{Statlog\_Gen} & \multicolumn{1}{|c|}{\textbf{0.63 $\pm$  0.02}} & \multicolumn{1}{|c|}{0.113  $\pm$  0.05} & \multicolumn{1}{|c|}{0.63  $\pm$  0.02} & \multicolumn{1}{|c|}{0.542  $\pm$  0.03} & \multicolumn{1}{|c|}{0.09  $\pm$  0.05} & \multicolumn{1}{|c|}{0.481  $\pm$  0.01} \\
    \multicolumn{1}{|c|}{12} & \multicolumn{1}{|c|}{Libras\_Gen} & \multicolumn{1}{|c|}{\textbf{0.913 $\pm$  0.048}} & \multicolumn{1}{|c|}{0.83 $\pm$  0.10} & \multicolumn{1}{|c|}{0.91 $\pm$  0.09} & \multicolumn{1}{|c|}{0.885 $\pm$  0.08} & \multicolumn{1}{|c|}{0.80 $\pm$  0.11} & \multicolumn{1}{|c|}{0.0 $\pm$  0.00} \\
    \multicolumn{1}{|c|}{13} & \multicolumn{1}{|c|}{Optical Digits\_Gen} & \multicolumn{1}{|c|}{\textbf{0.981 $\pm$  0.009}} & \multicolumn{1}{|c|}{0.828 $\pm$  0.018} & \multicolumn{1}{|c|}{0.981 $\pm$  0.009} & \multicolumn{1}{|c|}{0.952 $\pm$  0.019} & \multicolumn{1}{|c|}{0.826 $\pm$  0.009} & \multicolumn{1}{|c|}{0.839 $\pm$  0.03} \\
    \multicolumn{1}{|c|}{14} & \multicolumn{1}{|c|}{Ecoli\_Gen} & \multicolumn{1}{|c|}{\textbf{0.667 $\pm$  0.11}} & \multicolumn{1}{|c|}{0.540 $\pm$  0.14} & \multicolumn{1}{|c|}{0.64 $\pm$  0.16} & \multicolumn{1}{|c|}{0.690 $\pm$  0.13} & \multicolumn{1}{|c|}{0.587 $\pm$  0.09} & \multicolumn{1}{|c|}{0.61 $\pm$  0.17} \\
    \multicolumn{1}{|c|}{15} & \multicolumn{1}{|c|}{Car Evaluation\_Gen2} & \multicolumn{1}{|c|}{\textbf{0.955 $\pm$  0.01}} & \multicolumn{1}{|c|}{0.902 $\pm$  0.07} & \multicolumn{1}{|c|}{0.945 $\pm$  0.07} & \multicolumn{1}{|c|}{0.945 $\pm$  0.04} & \multicolumn{1}{|c|}{0.884 $\pm$  0.09} & \multicolumn{1}{|c|}{0.908 $\pm$  0.03} \\
    \multicolumn{1}{|c|}{16} & \multicolumn{1}{|c|}{US\_Crime\_Gen} & \multicolumn{1}{|c|}{\textbf{0.552 $\pm$  0.05}} & \multicolumn{1}{|c|}{0.50 $\pm$  0.04} & \multicolumn{1}{|c|}{0.51 $\pm$  0.06} & \multicolumn{1}{|c|}{0.53 $\pm$  0.03} & \multicolumn{1}{|c|}{0.497 $\pm$  0.1} & \multicolumn{1}{|c|}{0.511 $\pm$  0.08} \\
    \multicolumn{1}{|c|}{17} & \multicolumn{1}{|c|}{Protein\_homology} & \multicolumn{1}{|c|}{0.84 $\pm$  0.01} & \multicolumn{1}{|c|}{0.83 0.01} & \multicolumn{1}{|c|}{\textbf{0.86 $\pm$  0.01}} & \multicolumn{1}{|c|}{0.79 $\pm$  0.01} & -     & - \\
    \multicolumn{1}{|c|}{18} & \multicolumn{1}{|c|}{Scene\_Gen} & \multicolumn{1}{|c|}{\textbf{0.282 $\pm$  0.06}} & \multicolumn{1}{|c|}{0.061 $\pm$  0.09} & \multicolumn{1}{|c|}{0.25 $\pm$  0.06} & \multicolumn{1}{|c|}{0.190 $\pm$  0.08} & \multicolumn{1}{|c|}{0.131 $\pm$  0.09} & \multicolumn{1}{|c|}{0.15 $\pm$  0.001} \\
    \multicolumn{1}{|c|}{19} & \multicolumn{1}{|c|}{Solar\_Flare\_Gen} & \multicolumn{1}{|c|}{\textbf{0.262 $\pm$  0.09}} & \multicolumn{1}{|c|}{0  $\pm$  0} & \multicolumn{1}{|c|}{0  $\pm$  0} & \multicolumn{1}{|c|}{0.14  $\pm$  0.11} & \multicolumn{1}{|c|}{0  $\pm$  0} & \multicolumn{1}{|c|}{0  $\pm$  0} \\
    \hline
    \end{tabular}%
    }
  \label{tab:imbal_res_mcc}%
\end{table*}%

The p-values for the measures computed on the unbalanced datasets is shown in Table \ref{tab:res_pvals}. As these values are less that $0.05$, they do not indicate significant statistical difference.

\begin{table*}[htbp]
  \centering
  \caption{p-values for Wilcoxon signed ranks test for different measures on unbalanced Datasets}
    \begin{tabular}{|c|c|c|c|c|}
    \hline
        \textbf{S.No} & \textbf{Algorithm} & \textbf{G-Means} & \textbf{F-measure} & \textbf{MCC} \\
    \hline
    1     & Linear SVM & 1.32E-04 & 1.96E-04 & 1.31E-04  \\
    2     & Kernel SVM & 5.36E-04 & 1.17E-03 & 5.61E-03 \\
    3     & RFNN & 1.32E-04 & 1.36E-04 & 2.92E-04 \\
    \hline
    \end{tabular}%
  \label{tab:res_pvals}%
\end{table*}%

\begin{figure*}[hbtp]
    \centering
        \includegraphics[scale=0.5]{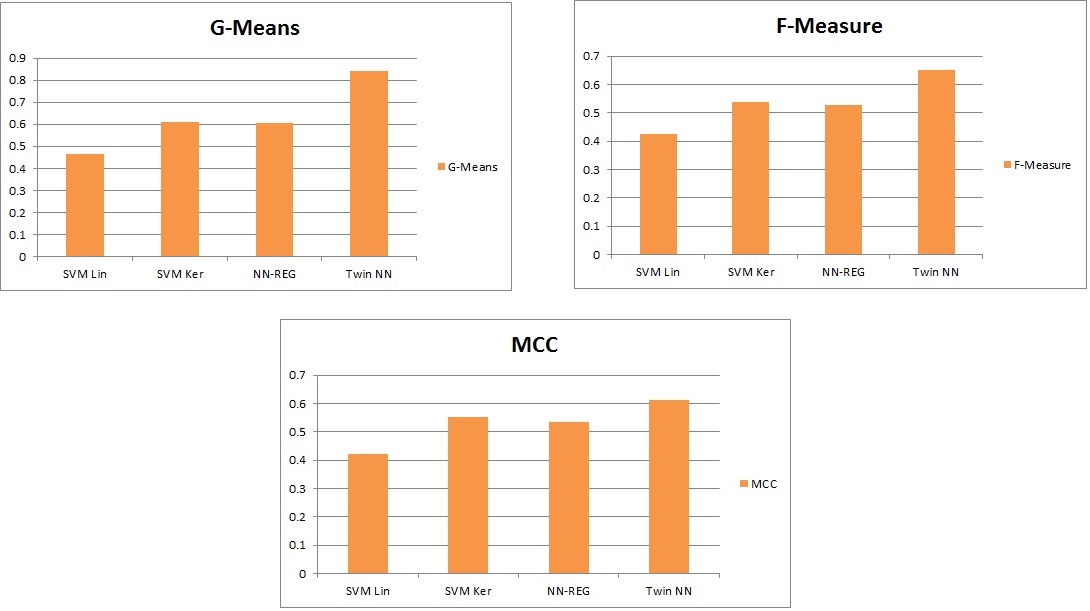}
        \caption{Average Values of G-Means, F-Measure, MCC and AUC for Unbalanced Datasets.}
        \label{fig:Avg_barchart}
\end{figure*}

We also show the average values of the measures computed on the unbalanced datasets in Table \ref{tab:res_imbal_avg_vals} for the various approaches. It can clearly be seen that the average values are highest for the Twin NN, indicating its superior performance across the unbalanced datasets. These are also graphically illustrated in Fig. \ref{fig:Avg_barchart}.

\begin{table}[htbp]
  \centering
  \caption{Average Values of G-Means, F-Measure, MCC and AUC for Unbalanced Datasets}
  \scalebox{0.75}{
    \begin{tabular}{|c|c|c|c|c|}
    \hline
    \textbf{Average} & \textbf{Linear SVM} & \textbf{Kernel SVM} & \textbf{RFNN} & \textbf{Twin NN} \\
    \hline
    G-Means & 0.465768 & 0.609421 & 0.607684 & 0.842195 \\
    F-Measure & 0.425158 & 0.539605 & 0.5272 & 0.651632 \\
    MCC   & 0.421547 & 0.551684 & 0.533789 & 0.612053 \\
    \hline
    \end{tabular}%
    }
  \label{tab:res_imbal_avg_vals}%
\end{table}%

Finally, to illustrate the scalability of the Twin NN on the unbalanced datasets, we present the training time for the Twin NN for these datasets in Table \ref{tab:imbal_res_time}. It can be seen that the Twin NN can be trained in tractable time for the unbalanced datasets.

\begin{table*}[htbp]
  \centering
  \caption{Training time for unbalanced datasets}
  \scalebox{0.7}{
    \begin{tabular}{|c|c|c|c|c|c|c|c|}
    \hline
    \textbf{S. No.} & \textbf{Dataset Generated} & \textbf{Twin NN} & \textbf{Lin SVM} & \textbf{Ker SVM} & \textbf{RFNN} & \textbf{Lin TWSVM} & \textbf{Ker TWSVM} \\
    \hline
    \multicolumn{1}{|c|}{\textbf{1}} & \multicolumn{1}{|c|}{\textbf{Abalone\_Gen1}} & \multicolumn{1}{|c|}{\textbf{0.292 $\pm$ 0.09}} & \multicolumn{1}{|c|}{0.55 $\pm$ 0.008} & \multicolumn{1}{|c|}{1.08 $\pm$ 0.02} & \multicolumn{1}{|c|}{1.05 $\pm$ 0.38} & \multicolumn{1}{|c|}{104 $\pm$ 7.58} & \multicolumn{1}{|c|}{231 $\pm$ 9.11} \\
    \multicolumn{1}{|c|}{\textbf{2}} & \multicolumn{1}{|c|}{\textbf{Letter\_Gen}} & \multicolumn{1}{|c|}{\textbf{5.01 $\pm$ 0.98}} & \multicolumn{1}{|c|}{15.01 $\pm$ 0.45} & \multicolumn{1}{|c|}{19.47 $\pm$ 0.49} & \multicolumn{1}{|c|}{6.88 $\pm$ 1.85} & -     & - \\
    \multicolumn{1}{|c|}{\textbf{3}} & \multicolumn{1}{|c|}{\textbf{Yeast\_Gen}} & \multicolumn{1}{|c|}{0.32 $\pm$ 0.1} & \multicolumn{1}{|c|}{\textbf{0.03 $\pm$ 0.0001}} & \multicolumn{1}{|c|}{0.09 $\pm$ 0.001} & \multicolumn{1}{|c|}{1.08$\pm$ 0.15} & \multicolumn{1}{|c|}{14.28 $\pm$ 0.53} & \multicolumn{1}{|c|}{35.27 $\pm$ 1.89} \\
    \multicolumn{1}{|c|}{\textbf{4}} & \multicolumn{1}{|c|}{\textbf{Abalone\_Gen2}} & \multicolumn{1}{|c|}{2.1 $\pm$ 0.1} & \multicolumn{1}{|c|}{\textbf{0.06 $\pm$ 0.002}} & \multicolumn{1}{|c|}{0.14 $\pm$ 0.001} & \multicolumn{1}{|c|}{1.7 $\pm$ 0.09} & \multicolumn{1}{|c|}{132 $\pm$ 13} & \multicolumn{1}{|c|}{255 $\pm$ 8.72} \\
    \multicolumn{1}{|c|}{\textbf{5}} & \multicolumn{1}{|c|}{\textbf{Coil\_2000}} & \multicolumn{1}{|c|}{\textbf{20.1 $\pm$ 2.2}} & \multicolumn{1}{|c|}{22.67 $\pm$ 0.18} & \multicolumn{1}{|c|}{27.06 $\pm$ 0.2} & \multicolumn{1}{|c|}{21.22 $\pm$ 4.21} & -     & - \\
    \multicolumn{1}{|c|}{\textbf{6}} & \multicolumn{1}{|c|}{\textbf{Car\_Eval\_Gen1}} & \multicolumn{1}{|c|}{0.8 $\pm$ 0.06} & \multicolumn{1}{|c|}{\textbf{0.07 $\pm$ 0.002}} & \multicolumn{1}{|c|}{0.5 $\pm$ 0.004} & \multicolumn{1}{|c|}{0.6 $\pm$ 0.009} & \multicolumn{1}{|c|}{8.36 $\pm$ 0.067} & \multicolumn{1}{|c|}{24.11 $\pm$ 1.17} \\
    \multicolumn{1}{|c|}{\textbf{7}} & \multicolumn{1}{|c|}{\textbf{Wine\_Quality\_Gen}} & \multicolumn{1}{|c|}{4.2 $\pm$ 0.8} & \multicolumn{1}{|c|}{\textbf{1.8 $\pm$ 0.007}} & \multicolumn{1}{|c|}{2.1 $\pm$ 0.05} & \multicolumn{1}{|c|}{2.1 $\pm$ 0.05} & \multicolumn{1}{|c|}{28.11$\pm$ 0.92} & \multicolumn{1}{|c|}{117.64$\pm$ 5.43} \\
    \multicolumn{1}{|c|}{\textbf{8}} & \multicolumn{1}{|c|}{\textbf{Ozone\_Level}} & \multicolumn{1}{|c|}{2.9 $\pm$ 0.06} & \multicolumn{1}{|c|}{\textbf{0.6 $\pm$ 0.02}} & \multicolumn{1}{|c|}{1.51 $\pm$ 0.008} & \multicolumn{1}{|c|}{1.98 $\pm$ 0.09} & \multicolumn{1}{|c|}{21.88 $\pm$ 2.7} & \multicolumn{1}{|c|}{114.5 $\pm$ 4.7} \\
    \multicolumn{1}{|c|}{\textbf{9}} & \multicolumn{1}{|c|}{\textbf{Pen Digits\_Gen}} & \multicolumn{1}{|c|}{\textbf{4.38$\pm$ 0.29}} & \multicolumn{1}{|c|}{6.79$\pm$ 0.29} & \multicolumn{1}{|c|}{9.36$\pm$ 0.48} & \multicolumn{1}{|c|}{4.68$\pm$ 0.12} & -     & - \\
    \multicolumn{1}{|c|}{\textbf{10}} & \multicolumn{1}{|c|}{\textbf{Spectrometer\_Gen}} & \multicolumn{1}{|c|}{0.31$\pm$ 0.03} & \multicolumn{1}{|c|}{\textbf{0.09$\pm$ 0.008}} & \multicolumn{1}{|c|}{0.10$\pm$ 0.003} & \multicolumn{1}{|c|}{0.58$\pm$ 0.18} & \multicolumn{1}{|c|}{0.60$\pm$ 0.07} & \multicolumn{1}{|c|}{4.81$\pm$ 0.21} \\
    \multicolumn{1}{|c|}{\textbf{11}} & \multicolumn{1}{|c|}{\textbf{Statlog\_Gen}} & \multicolumn{1}{|c|}{\textbf{5.21$\pm$ 0.9}} & \multicolumn{1}{|c|}{5.35$\pm$ 0.1} & \multicolumn{1}{|c|}{5.61$\pm$ 0.5} & \multicolumn{1}{|c|}{5.91$\pm$ 0.5} & \multicolumn{1}{|c|}{55.97$\pm$ 1.87} & \multicolumn{1}{|c|}{127.97$\pm$ 2.88} \\
    \multicolumn{1}{|c|}{\textbf{12}} & \multicolumn{1}{|c|}{\textbf{Libras\_Gen}} & \multicolumn{1}{|c|}{\textbf{0.50$\pm$ 0.007}} & \multicolumn{1}{|c|}{0.69$\pm$ 0.011} & \multicolumn{1}{|c|}{0.75$\pm$ 0.02} & \multicolumn{1}{|c|}{0.63$\pm$ 0.11} & \multicolumn{1}{|c|}{0.24$\pm$ 0.008} & \multicolumn{1}{|c|}{1.13$\pm$ 0.01} \\
    \multicolumn{1}{|c|}{\textbf{13}} & \multicolumn{1}{|c|}{\textbf{Optical Digits\_Gen}} & \multicolumn{1}{|c|}{\textbf{9.18$\pm$ 0.30}} & \multicolumn{1}{|c|}{10.30$\pm$ 005} & \multicolumn{1}{|c|}{20.53$\pm$ 0.17} & \multicolumn{1}{|c|}{9.87$\pm$ 0.33} & \multicolumn{1}{|c|}{65.72$\pm$ 1.79} & \multicolumn{1}{|c|}{132.72$\pm$ 3.21} \\
    \multicolumn{1}{|c|}{\textbf{14}} & \multicolumn{1}{|c|}{\textbf{Ecoli\_Gen}} & \multicolumn{1}{|c|}{0.15$\pm$ 0.01} & \multicolumn{1}{|c|}{\textbf{0.01$\pm$ 0.0008}} & \multicolumn{1}{|c|}{0.05$\pm$ 0.001} & \multicolumn{1}{|c|}{0.40$\pm$ 0.1} & \multicolumn{1}{|c|}{0.21$\pm$ 0.001} & \multicolumn{1}{|c|}{1.08$\pm$ 0.08} \\
    \multicolumn{1}{|c|}{\textbf{15}} & \multicolumn{1}{|c|}{\textbf{Car Evaluation\_Gen2}} & \multicolumn{1}{|c|}{0.61$\pm$ 0.01} & \multicolumn{1}{|c|}{\textbf{0.13$\pm$ 0.005}} & \multicolumn{1}{|c|}{0.60$\pm$ 0.009} & \multicolumn{1}{|c|}{0.59$\pm$ 0.05} & \multicolumn{1}{|c|}{11.85$\pm$ 0.02} & \multicolumn{1}{|c|}{50.34$\pm$ 1.75} \\
    \multicolumn{1}{|c|}{\textbf{16}} & \multicolumn{1}{|c|}{\textbf{US\_Crime\_Gen}} & \multicolumn{1}{|c|}{3.8$\pm$ 0.23} & \multicolumn{1}{|c|}{\textbf{0.89$\pm$ 0.03}} & \multicolumn{1}{|c|}{2.05$\pm$ 0.02} & \multicolumn{1}{|c|}{1.84$\pm$ 0.27} & \multicolumn{1}{|c|}{27.83$\pm$ 7.82} & \multicolumn{1}{|c|}{56.87$\pm$ 0.91} \\
    \multicolumn{1}{|c|}{\textbf{17}} & \multicolumn{1}{|c|}{\textbf{Protein\_homology}} & \multicolumn{1}{|c|}{\textbf{132.2$\pm$ 2.87}} & \multicolumn{1}{|c|}{235.12$\pm$ 8.19} & \multicolumn{1}{|c|}{744.78$\pm$ 32.71} & \multicolumn{1}{|c|}{188.2$\pm$5.22} & -     & - \\
    \multicolumn{1}{|c|}{\textbf{18}} & \multicolumn{1}{|c|}{\textbf{Scene\_Gen}} & \multicolumn{1}{|c|}{\textbf{6.2$\pm$ 1.01}} & \multicolumn{1}{|c|}{6.6$\pm$ 0.11} & \multicolumn{1}{|c|}{11.9$\pm$ 0.09} & \multicolumn{1}{|c|}{10.03$\pm$ 0.24} & \multicolumn{1}{|c|}{32.60$\pm$ 0.79} & \multicolumn{1}{|c|}{137.46$\pm$ 7.48} \\
    \multicolumn{1}{|c|}{\textbf{19}} & \multicolumn{1}{|c|}{\textbf{Solar\_Flare\_Gen}} & \multicolumn{1}{|c|}{0.20$\pm$ 0.03} & \multicolumn{1}{|c|}{\textbf{0.10$\pm$ 0.006}} & \multicolumn{1}{|c|}{0.23$\pm$ 0.008} & \multicolumn{1}{|c|}{0.63$\pm$ 0.07} & \multicolumn{1}{|c|}{3.77$\pm$ 0.6} & \multicolumn{1}{|c|}{8.7$\pm$ 1.1} \\
    \hline
    \end{tabular}%
    }
  \label{tab:imbal_res_time}%
\end{table*}%

\subsection{Experiments on scalability}

We analyze the scalability of the Twin NN on the Forest Covertype dataset described in Table \ref{tab:scaling_dataset_description}. The dataset with imbalance generated has an imbalance ratio of $1:211$, with 44 numeric and 10 categorical attributes. 

\begin{table}[h]
  \centering
  \caption{Dataset for scaling experiment}
  \scalebox{0.75}{
    \begin{tabular}{|c|c|}
    \hline
    Dataset Generated & Forest\_CovType\_Gen \\
  \hline    Source & UCI KDD \\
  \hline    Area  & Nature \\
  \hline    Imbalance Ratio & 1:211 \\
  \hline    Number of Samples per class & 2747 : 578,265 \\
  \hline    Attributes & 44N,10C \\
  \hline    Original Dataset & Forest\_CovType \\
  \hline    Original Classes & 7 \\
  \hline    Class No. used to generate imbalance & 4 \\
    \hline
    \end{tabular}%
    }
  \label{tab:scaling_dataset_description}%
\end{table}%

\begin{figure*}[hbtp]
    \centering
        \includegraphics[scale=0.45]{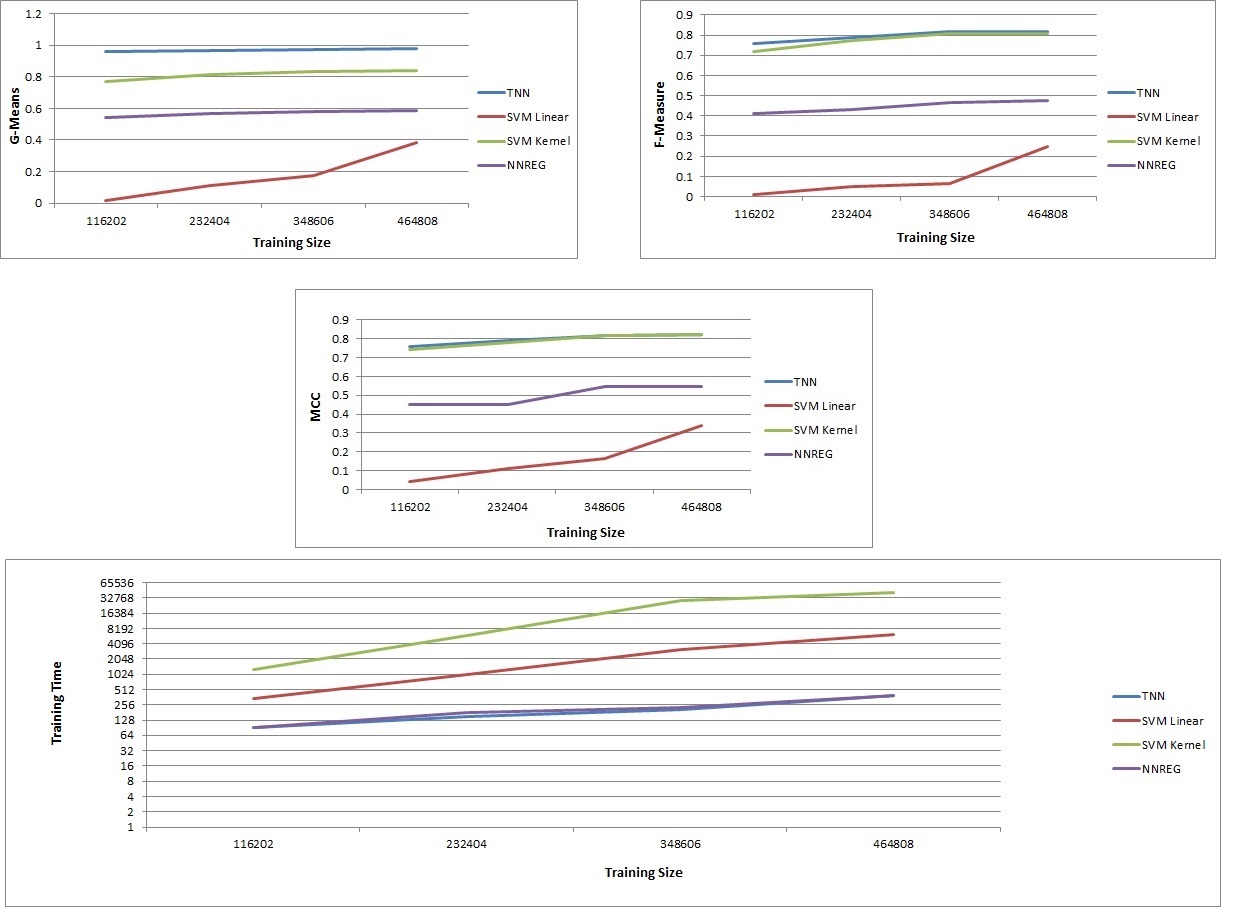}
        \caption{Average Test  Values of G-Means, F-Measure, MCC and AUC and Training Time for Scale up Experiment. }
        \label{fig:Avg_barchart_scaling}
\end{figure*}

The performance parameters for this case have been shown in Table \ref{tab:scaling_gmeans} for various number of training samples, and the corresponding training time has been shown in Table \ref{tab:scaling_time}. These are also graphically illustrated in Fig. \ref{fig:Avg_barchart_scaling}. It can be seen that in comparison to the other methods, the Twin NN scales well and also trains in tractable time with increase in number of training samples. This establishes the scalability of the Twin NN for imbalanced datasets.

\begin{table}[h]
  \centering
  \caption{Performance parameters for scaling experiment }
  \scalebox{0.75}{
    \begin{tabular}{|c|c|c|c|c|}
    \hline
    
    \textbf{Train Data Size} & TNN   & SVM Linear  & SVM Kernel & NNREG \\
    \hline
    \hline
    \multicolumn{5}{|c|}{G-means values}\\
    \hline
        \hline
    116202 & 0.9627 & 0.017 & 0.7712 & 0.54 \\
    232404 & 0.9689 & 0.112 & 0.8132 & 0.57 \\
    348606 & 0.9731 & 0.1754 & 0.8361 & 0.5808 \\
    464808 & 0.9802 & 0.3857 & 0.8391 & 0.5885 \\
    \hline
    \hline
    \multicolumn{5}{|c|}{F-measure}\\
    \hline
        \hline
       116202 & 0.7581 & 0.01  & 0.72  & 0.41 \\
    232404 & 0.7881 & 0.051 & 0.771 & 0.43 \\
    348606 & 0.8172 & 0.0653 & 0.8069 & 0.467 \\
    464808 & 0.8172 & 0.2485 & 0.8071 & 0.4736 \\
 \hline    
    \hline
    \multicolumn{5}{|c|}{MCC values}\\
   \hline
       \hline
   116202 & 0.7611 & 0.041 & 0.74  & 0.4523 \\
    232404 & 0.7911 & 0.112 & 0.78  & 0.4512 \\
    348606 & 0.8162 & 0.1649 & 0.8154 & 0.5464 \\
    464808 & 0.8198 & 0.3382 & 0.8196 & 0.5487 \\
 \hline     
    \hline
    \end{tabular}%
    }
  \label{tab:scaling_gmeans}%
\end{table}%

\begin{table}[H]
  \centering
  \caption{Training Time  for scaling experiment }
  \scalebox{0.75}{
    \begin{tabular}{|c|c|c|c|c|}
    \hline
         
    \hline
    \textbf{Train Data Size} & TNN   & SVM Linear  & SVM Kernel & NNREG \\
    \hline   
    116202 & 89.48 & 338.8 & 1290.11 & 91.11 \\
    232404 & 149.38 & 990.4 & 5811.32 & 181.27 \\
    348606 & 209.41 & 3129.21 & 28716.11 & 227.41 \\
    464808 & 385.93 & 6254.1 & 41861.21 & 394.71 \\
    \hline
    \end{tabular}%
    }
  \label{tab:scaling_time}%
\end{table}%

\subsection{Results on multi-class datasets}

We evaluate the performance of the multi-class architecture of the Twin NN on a few multi-class datasets. A summary of the datasets is shown in Table \ref{tab:largedatasets}. For datasets where training, validation and test sets have been explicitly provided, we have used the same for evaluation. For other cases, we have employed the same procedure as for the UCI datasets to split the data into train, validation or test sets and those entries are marked by a `-' in Table \ref{tab:largedatasets}.

\begin{table}[htbp]
	\centering
	\caption{Multi-class datasets used in experiments}
	\begin{tabular}{|l|c|c|c|c|c|}
		\hline
		\multicolumn{1}{|l|}{Dataset}& \multicolumn{1}{|l|}{\# train} & \multicolumn{1}{|c|}{\# val} & \multicolumn{1}{|c|}{\# test} & \multicolumn{1}{|c|}{features} & \multicolumn{1}{|c|}{\# classes} \\
		\hline
		\hline acoustic   & 78823 & -  & 19705  & 50   & 3 \\
		\hline connect4 & 67757 & - & - & 126     & 3 \\
		\hline pendigits  & 7494  & -  & 3498 & 16   & 10 \\
		
		\hline satimage & 3104  & 1331  & 2000  & 36    & 6 \\
		\hline segment & 1386  & 462   & 462   & 19    & 7 \\
		\hline shuttle  & 30450  & 13050  & 14500 & 9   & 7 \\
		\hline dna   & 1400  & 600   & 1186  & 180   & 3 \\		
		\hline letter & 10500 & 4500  & 5000  & 16    & 26 \\
		\hline seismic   & 78823 & -  & 19705 & 50   & 3 \\
		\hline mnist & 47999 & 12001 & 10000 & 778   & 10 \\
		\hline
	\end{tabular}%
	\label{tab:largedatasets}%
\end{table}%

The performance of the Twin NN on the multi-class datasets is shown in Table \ref{tab:mc_results}. It can be seen that the Twin NN performs better than SVM (with RBF kernel), neural network with regularization (NN) and Random Forests (RF) for most of the datasets.

\begin{table}[hbtp]
\centering
\caption{Results on multi-class datasets}
\label{tab:mc_results}
\begin{tabular}{|c|c|c|c|c|c|}
\hline
S. No. & Dataset   & Twin NN & SVM RBF & NN    & RF    \\
\hline
\hline 1      & acoustic  & \textbf{79.58}   & 75.29   & 78.39 & 79.3  \\
\hline 2      & connect4  & \textbf{83.95}   & 83.65   & 80.89 & 82.14 \\
\hline 3      & pendigits & 99.14   & 99.60   & 99.59 & 99.06 \\
\hline 4      & satimage  & \textbf{91.32}   & 89.7    & 89.17 & 90.3  \\
\hline 5      & segment   & \textbf{95.46}   & 95      & 92.64 & 93.72 \\
\hline 6      & shuttle   & 99.37   & 99.61   & 98.53 & \textbf{99.98} \\
\hline 7      & dna       & \textbf{95.37}   &   94.18      & 93.66 & 92.66 \\
\hline 8      & letter    &   \textbf{96.99}      & 96.2    & 90.04      & 94.9  \\
\hline 9      & seismic    & \textbf{76.58}   & 72.61   & 76.1  & 75.5  \\
\hline 10     & mnist     & \textbf{98.13}   & 97.02   & 96.25 & 96.68  \\
\hline  
\end{tabular}
\end{table}

\subsubsection{Handling high imbalance in multi-class datasets}

\begin{figure}[hbtp]
\centering
\begin{subfigure}[b]{0.45\textwidth}
 \includegraphics[width=\textwidth]{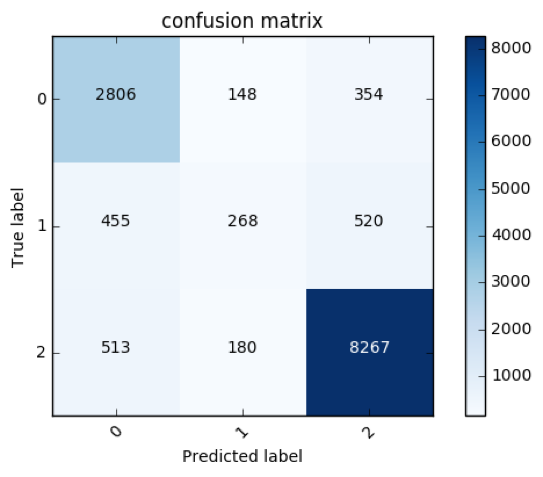}
 \caption{Twin NN, test acc=83.95\%}
 \label{fig:twin_nn_acc}
\end{subfigure}
\quad
\begin{subfigure}[b]{0.45\textwidth}
 \includegraphics[width=\textwidth]{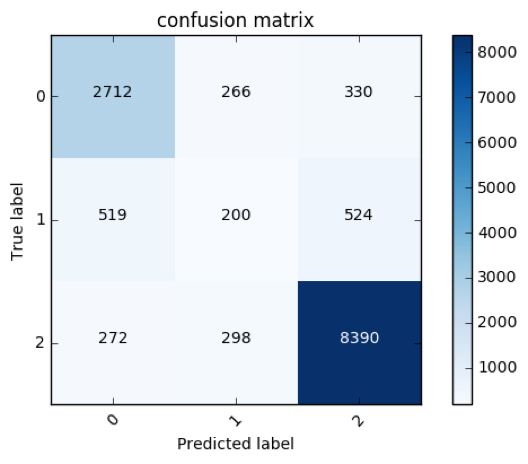}
 \caption{SVM with RBF kernel, test acc=83.65\%}
 \label{fig:svm_rbf}
\end{subfigure}
\quad
\begin{subfigure}[b]{0.45\textwidth}
 \includegraphics[width=\textwidth]{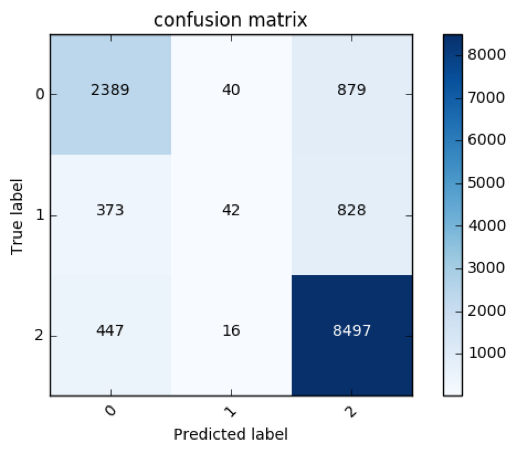}
 \caption{Neural Network with Regularization, test acc=80.89\%}
 \label{fig:nn}
\end{subfigure}
\quad
\begin{subfigure}[b]{0.45\textwidth}
 \includegraphics[width=\textwidth]{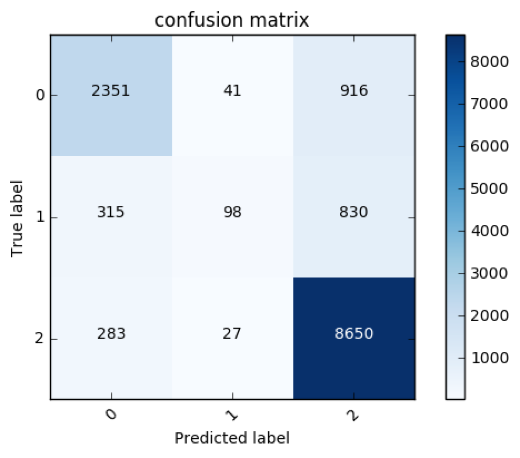}
 \caption{Random Forest, test acc=82.14\%}
 \label{fig:rf}
\end{subfigure}
\caption{Confusion matrices for different models for the Connect4 dataset.}
\label{fig:connect4}
\end{figure}

We also illustrate the benefit of using the Twin NN in cases where there is high imbalance for a multi-class dataset. We consider the Connect4 dataset, which has 3 classes and 126 features. The number of the samples in the 3 classes in the training set are 11,654, 4551 and 31,086, which indicates high skewness in the distribution of samples between the classes. Figure \ref{fig:connect4} shows the confusion matrix for the testing samples of this dataset using different learning models.

It can be observed that the Twin NN correctly classifies more samples of the minority class when compared to competing methods (268 in comparison to 200, 42 and 98 for SVM, neural networks and random forest respectively), while also giving the highest accuracy on the dataset. This illustrates that the Twin NN is beneficial even in cases of imbalance in multi-class datasets.

\section{Conclusion and Future Work \label{sec:conc}}
In this paper, we present the Twin Neural Network, which extends the motivation and ideas behind the Twin SVM into a novel neural network framework. In the case of binary classification, the Twin NN tries to learn a hyperplane passing through samples of one class but at a distance of at least one from samples of the other class. The training targets for the Twin NN are very different from a conventional neural network. The class of an output sample is determined by using the distances from the hyperplanes associated with each class. We also present a multiclass extension of the Twin NN, in which multiple hyperplanes are associated with each class. The Twin NN has the advantage that the feature map for each class is separately optimized. The Twin NN has been shown to have superior performance on several datasets, particularly for the case of unbalanced datasets. The Twin NN architecture has also been extended for classification of multi-class datasets and has shown to give good performance on such datasets as well. It is also substantially faster. Future work would involve developing architectures for other formulations based on the TWSVM, as well as work on variants for regression.

\section*{Acknowledgements}
The corresponding author would like to acknowledge the support of the Microsoft Chair Professor Project Grant (MI01158, IIT Delhi). 
\bibliographystyle{plain}
\bibliography{ref}

\end{document}